\documentclass[letterpaper]{article} 
\usepackage[]{aaai2026}  
\usepackage{times}  
\usepackage{helvet}  
\usepackage{courier}  
\usepackage{amssymb}
\usepackage[hyphens]{url}  
\usepackage{graphicx} 
\usepackage{amsthm}
\newtheorem{definition}{Definition}
\newtheorem{property}{Property}
\usepackage{natbib}  
\usepackage{caption} 
\usepackage{algorithm}
\usepackage{algorithmic}

\usepackage{booktabs}
\usepackage{multirow}
\usepackage{makecell}
\usepackage{subcaption}
\usepackage{amsmath}
\usepackage{xcolor}
\usepackage{longtable}  
\usepackage{tabularx}  
\usepackage{tikz} 
\usetikzlibrary{shapes, arrows, fit, positioning} 

\newcommand{\codeurl}[1]{\url{#1}}
\definecolor{elem1color}{HTML}{0077BB} 
\definecolor{elem2color}{HTML}{D9534F} 
\definecolor{elem3color}{HTML}{993399} 
\definecolor{elem4color}{HTML}{2A8000} 
\newcommand{\std}[1]{\,\scriptsize{(#1)}}
\definecolor{goodgreen}{HTML}{2A8000}
\definecolor{badred}{HTML}{D9534F}

\definecolor{steelblue}{rgb}{0.27, 0.51, 0.7}
\definecolor{azure}{rgb}{0.0, 0.5, 1.0}
\definecolor{nicegreen}{rgb}{0.0, 0.7, 0.1}
\definecolor{CuGray}{gray}{0.9}
\definecolor{amethyst}{rgb}{0.6, 0.4, 0.8}
\definecolor{black}{rgb}{0.0, 0.0, 0.0}
\newcommand{\mk}[1]{\textcolor{black}{#1}}

\newcommand{\styleEone}[1]{\textbf{\textcolor{elem1color}{#1}}}
\newcommand{\styleEtwo}[1]{\textbf{\textcolor{elem2color}{#1}}}
\newcommand{\styleEthree}[1]{\textbf{\textcolor{elem3color}{#1}}}
\newcommand{\styleEfour}[1]{\textbf{\textcolor{elem4color}{#1}}}

\usepackage{newfloat}
\usepackage{listings}
\DeclareCaptionStyle{ruled}{labelfont=normalfont,labelsep=colon,strut=off} 
\floatstyle{ruled}
\newfloat{listing}{tb}{lst}{}
\floatname{listing}{Listing}
\title{
Aligning Large Language Models with Procedural Rules:\\
An Autoregressive State-Tracking Prompting for In-Game Trading
} 

\author{
}

\author {
    Minkyung Kim\textsuperscript{\rm 1},
    Junsik Kim\textsuperscript{\rm 2},
    Woongcheol Yang\textsuperscript{\rm 1},
    Sangdon Park\textsuperscript{\rm 1}\textsuperscript{$\dagger$},
    Sohee Bae\textsuperscript{\rm 1}\textsuperscript{$\dagger$}
}
\affiliations {
    \textsuperscript{\rm 1}SayBerryGames,
    \textsuperscript{\rm 2}Amazon\\
    \textsuperscript{$\dagger$}Co-corresponding authors\\
    mkkim1778@sayberrygames.com, jskimcv@amazon.com, woongy@sayberrygames.com,\\ 
    chaos@sayberrygames.com, bread@sayberrygames.com
}

\usepackage{bibentry}

\nocopyright

\begin{document}

\maketitle

\begingroup 
\renewcommand\thefootnote{}
\footnotetext{© 2025 Copyright held by the owner/author(s).} 
\endgroup 

\begin{abstract}
Large Language Models (LLMs) enable dynamic game interactions but fail to follow essential procedural flows in rule-governed trading systems, eroding player trust. 
\mk{This work} resolves the core tension between the creative flexibility of LLMs and \mk{the procedural demands of in-game trading}
(browse-offer-review-confirm).
\mk{To this end,} Autoregressive State-Tracking Prompting (ASTP) is introduced, \mk{a methodology centered on a strategically orchestrated prompt that compels an LLM to make its state-tracking process explicit and verifiable. Instead of relying on implicit contextual understanding, ASTP tasks the LLM with identifying and reporting a predefined state label from the previous turn. To ensure transactional integrity, this is complemented by a state-specific placeholder post-processing method for accurate price calculations.} 
Evaluation across 300 trading dialogues demonstrates $>$99\% state compliance
and 99.3\% calculation precision. Notably, ASTP with placeholder post-processing on smaller models (Gemini-2.5-Flash) matches larger models' (Gemini-2.5-Pro) performance while reducing response time from 21.2s to 2.4s, establishing a practical foundation that satisfies both real-time requirements and resource constraints of commercial games.
\textit{The code will be publicly available upon acceptance.}
\end{abstract}


\section{Introduction}

\begin{figure*}[t!]
    \centering
    \includegraphics[width=0.98\linewidth]{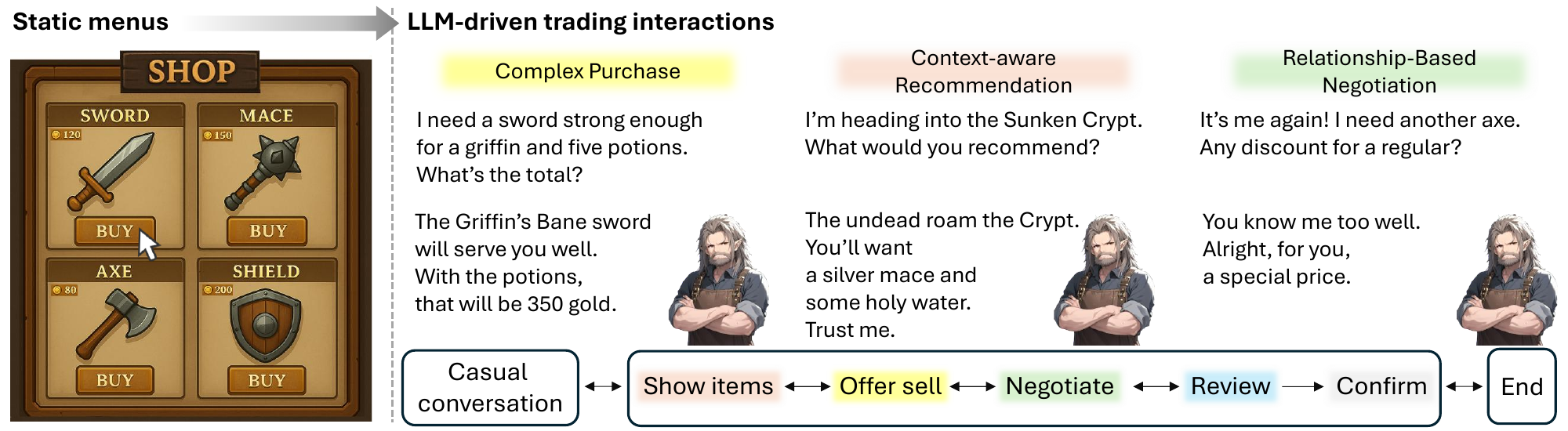}
    \caption{Comparison of static menus and LLM-driven trading interactions showcasing complex purchases, context-aware recommendations, and relationship-based negotiations within a semi-structured dialogue flow.}
    \label{fig:teaser_concept}
\end{figure*}

Large Language Models (LLMs) are transforming non-player character (NPC) interactions in games, enabling natural language-driven experiences that transcend traditional script constraints \citep{wang2024survey, maleki2024procedural, ReluGames2025SmokingGun, Krafton2025InZoi}. A critical yet underexplored application for these dynamic NPCs is in-game trading, a core mechanic connecting economic activities and character development\cite{kim2024mart}. While LLMs promise to replace static menus with emergent gameplay, enabling context-aware recommendations and relationship-based negotiations (Figure~\ref{fig:teaser_concept}), their deployment faces a fundamental challenge.

The core tension lies in reconciling LLMs' creative flexibility with the semi-structured procedures of commercial transactions. In-game trading must follow a flow of browse, offer, review, and confirm, serving as an essential safeguard to protect player assets. 
\mk{Yet, the unconstrained, intent-driven nature of LLMs makes them prone to circumventing this structured flow\cite{agrawal2024dfi, kodama2023metacontrol, wu2024stateflow}. For instance, their tendency to directly satisfy a perceived user goal can lead}
a simple price inquiry to trigger an unwanted purchase, or a negotiation skipping the critical review step. Such procedural violations erode player trust and undermine system integrity.

This challenge extends beyond gaming to any rule-governed human-AI interaction requiring both conversational naturalness and procedural compliance, such as in customer service protocols or medical consultations. Despite its importance, existing approaches either sacrifice flexibility for structure\cite{kodama2023metacontrol} or fail to enforce critical procedural constraints, leaving a significant gap in deploying LLMs for real-world applications where both qualities are essential.

To address this, we introduce Autoregressive State-Tracking Prompting (ASTP), a prompting methodology that resolves the tension between flexibility and control by making state-tracking an explicit, autoregressive process. Instead of relying on general reasoning heuristics (e.g., ``think step-by-step"), ASTP requires the LLM to infer and output the previous dialogue state's label before predicting the next. This verifiable process is embedded in a structured Prime–Guide–Enforce workflow that maintains procedural compliance without compromising conversational naturalness. Additionally, we demonstrate that this state-aware architecture enables a practical post-processing mechanism that boosts numerical reliability, 
allowing smaller models to match the accuracy of larger ones with substantial speed gains.

Our main contributions are as follows:
\begin{itemize}
    \item We propose a prompting strategy, \textbf{Autoregressive State-Tracking Prompting (ASTP)}, that makes dialogue state transitions explicit and verifiable.
    \item We design a structured prompt workflow, \textbf{Prime–Guide–Enforce}, that significantly improves procedural compliance, increasing adherence to key safeguards from 78.1\% to 99.6\%.
    \item We develop a placeholder-based post-processing method, \textbf{PPP}, that improves \mk{price calculation} accuracy in trading tasks from 84.3\% to 99.3\%, enabling smaller models to match the accuracy of larger ones.

\end{itemize}


\section{Related Work}

One of the well-established fields for managing goal-oriented dialogue interactions is Task-Oriented Dialogue (TOD). TOD aims to help a user complete a specific task, e.g., booking a flight. A traditional TOD system relies on core components like a dialogue state tracker, which fills predefined `slots' (e.g., destination, time) from user utterances\cite{feng2023ldst, hudecek2023llmtod, li2024fnctod}, and a dialogue policy, which decides the next system action (e.g., calling API or requesting additional information) based on the current state\cite{Schick2023ToolUse, cao2024diaggpt}.

The paradigm has shifted towards end-to-end systems that replace this modular pipeline\cite{hosseini2020simple, saley2024synctod}. AutoTOD\cite{xu2024autotod}, specifically, re-architects the workflow using a single unified prompt that decomposes a complex goal into a series of `sub-tasks'. The system then selects the most appropriate sub-task for a user's intent in each turn and executes the prompt for each sub-task, guiding the LLM to implicitly handle state-tracking and policy decisions. This highlights a fundamental characteristic of the TOD paradigm: its dialogue policy is optimized to achieve a goal, not to enforce a specific sequence of interactions. This goal-driven optimization, while effective for flexible task handling, is fundamentally misaligned with the demand for strict procedural adherence, as it makes the system prone to bypassing critical safeguards like a final review.

Another approach to governing LLM interactions involves explicitly defining the conversational path. Recognizing that open-ended generation is often insufficient for goal-oriented tasks, this line of research focuses on providing a blueprint to guide the LLM\cite{agrawal2024dfi, du2024dflow}. However, while these methods excel at the design of the flow, the challenge of developing a robust runtime mechanism to compel the LLM to adhere to it remains underexplored.
Distinct from defining a conversational path, other methods model the workflow based on the task-solving state, 
enhancing control and efficiency\cite{wu2024stateflow, rozanov2024stateact}.

Prompting techniques that enhance an LLM's reasoning, such as Chain-of-Thought (CoT), have proven remarkably effective at unlocking complex reasoning capabilities in large models\cite{wei2022chain, kojima2022large, wang2022self, Bai2022ConstitutionalAI, yao2023tree, madaan2023self}. These methods typically prompt the LLM to ``think step-by-step" before responding, which can improve performance on a wide range of tasks. However, their reasoning process is often unconstrained and generic. In contrast, for applications requiring procedural adherence, explicitly specifying what the LLM should think about offers a more robust approach. Our approach transforms vague 'thinking' into a 'verifiable procedure' by instructing the model to first identify the previous state and then compelling it to report that state in its response. This targeted approach provides a more robust anchor for the model's reasoning, making it more suitable for procedural tasks than generic instructions.

Meanwhile, research in our direct application domain, game NPCs, has focused on a different trajectory. Generative Agents\cite{park2023generative} aim to create believable
NPCs through sophisticated memory mechanisms, while recent work on persistence\cite{song2025llm} and narrative generation\cite{peng2024player} pushes creative boundaries.
Yet these systems deliberately avoid structured constraints, prioritizing emergent behavior\cite{lan2024avalon} over reliability. Even MART\cite{kim2024mart}, which specifically targets merchant NPCs, reveals this fundamental gap. Specifically, it handles price negotiation but cannot manage multi-item transactions or enforce the purchase confirmations essential for protecting player assets.

In summary, the landscape of related work reveals a clear and critical gap. Task-oriented systems prioritize goal completion over procedural flow, design-focused frameworks lack enforcement mechanisms, general reasoning techniques are not tailored for strict procedural control, and research in game NPCs has favored creativity over compliance. This highlights the need for a new approach specifically designed to ensure robust, verifiable procedural adherence within conversational agents. Our work aims to fill this gap.

\section{Proposed Method}


This section details a method for building a reliable merchant NPC by addressing both conversational coherence and transactional integrity. The method first introduces Autoregressive State-Tracking Prompting (ASTP) to enforce a logical dialogue flow, and then integrates a state-specific post-processing rule as a vital component to guarantee computational accuracy within that flow.

\subsection{Prompt Architecture}
In the game, players can naturally converse and trade with merchant NPCs. The prompt for the merchant NPC is structured to provide rich contextual information for the LLM. While the prompt contains several components, the core of the proposed method lies in the design of the Dialogue Guidelines within a unified prompt. The full prompt consists of the following.
\begin{itemize}
    \item \textbf{System Instructions}: Defines the NPC's persona, role, and behavioral directives. It also includes situational context like the in-game location and time to ground its response in the game world.
    \item \textbf{Game World Data}: Contains two distinct sets of item data: a list of all items existing in the game world, and a separate list of the NPC's actual sellable inventory. This dual structure allows the NPC to act as both an immersive world informant and a reliable merchant.
    \item \textbf{Dialogue History}: Presents preceding conversation turns, including the player's latest utterance and the context inferred by the NPC from previous interactions.
    \item \textbf{Dialogue Guidelines}: Provides the core logic for dialogue state management. These rules instruct the LLM on how to infer the current dialogue state and when to transition between states, which will be detailed in the following sections.
    \item \textbf{Expected Response Format}: Instructs the LLM to generate its output as a single, well-formed JSON object. This approach relies on the LLM's capability to follow formatting instructions within the prompt, rather than utilizing specialized structured output functionalities. Key fields in the JSON object include NPC's spoken dialogue, the inferred dialogue state, and data pertinent to the current transaction, which contains information such as an array of items with their quantities and prices.
\end{itemize}

\subsection{ASTP: Formalization and Principles}

\subsubsection{State Space}
We define the system's dialogue state space $\mathcal{S}$ as follows:
\begin{itemize}
    \item \textbf{Base state set ($S_{\text{base}}$):} Main dialogue states unrelated to trading.
    $$S_{\text{base}} = \{\texttt{CASUAL}, \texttt{END}\}$$
    \item \textbf{Trade state set ($S_{\text{trade}}$):} Specific states possible within a trading context.
    \begin{align*}
    S_{\text{trade}} = \{&\texttt{SHOW\_ITEMS}, \texttt{OFFER\_SELL}, \texttt{NEGOTIATE},\\
    &\texttt{FINAL\_CHECK}, \texttt{COMMIT\_SALE}\}
    \end{align*}
    \item \textbf{Complete state space ($\mathcal{S}$):} The union of both sets.
    $$\mathcal{S} = S_{\text{base}} \cup S_{\text{trade}}$$
\end{itemize}

\subsubsection{Dialogue History}
The dialogue history $h_t$ up to turn $t$ is defined as a sequence of user utterances ($u$) and system responses ($r$):
$$h_t = (u_1, r_1, u_2, r_2, ..., u_t, r_t)$$

We define a state extraction function to retrieve the previous state from the last response:
$$\textit{last\_state}(h_t) = \begin{cases}
s \in S_{\text{trade}} & \text{if in a trading context,} \\
s \in S_{\text{base}} & \text{otherwise.}
\end{cases}$$


\subsubsection{ASTP Function}
The core functionality of Autoregressive State-Tracking Prompting (ASTP) is modeled as:

\begin{definition}[ASTP Function] \label{def:astp}
Given $(h_t, u_{t+1}, \mathcal{G})$, 
the Autoregressive State-Tracking Prompting function 
$$f_{\text{ASTP}}: \mathcal{H} \times \mathcal{U} \times \mathcal{G} \mapsto \mathcal{S} \times \mathcal{S} \times \mathcal{R}$$
computes $(s_t, s_{t+1}, r_{t+1})$ through the following sequential process:
\begin{enumerate}
    \item \textbf{State Inference:} Infers the previous dialogue state from the conversation history
    $$s_t = \text{infer\_previous\_state}(h_t)$$
    
    \item \textbf{State Transition:} Determines the current state based on the inferred previous state and user input
    $$s_{t+1} = \text{determine\_current\_state}(s_t, u_{t+1}, \mathcal{G})$$
    
    \item \textbf{Response Generation:} Generates an appropriate NPC response for the determined state
    $$r_{t+1} = \text{generate\_response}(s_{t+1}, u_{t+1}, \mathcal{G})$$
\end{enumerate}

\noindent where, $\mathcal{H}, \mathcal{U}, \mathcal{G}, \mathcal{R}$ denote the spaces of dialogue histories, user utterances, game contexts included in the prompt, and system responses, respectively.
\end{definition}
The key insight is that $s_t$ is not directly extracted but inferred by the LLM from the dialogue history, enabling state-aware decision making.

\subsubsection{State Transition Constraints}
While the ASTP function autonomously infers states, it must satisfy certain transition constraints for system stability and rule compliance. Let $\mathcal{E} \subseteq \mathcal{S} \times \mathcal{S}$ denote the set of valid state transitions permitted by the system.

\begin{property}[Valid ASTP Execution]
Any valid execution of $f_{\text{ASTP}}$ must satisfy:
\begin{enumerate}
    \item \textbf{Transition Validity:} $(s_t, s_{t+1}) \in \mathcal{E}$
    
    \item \textbf{Critical Transaction Constraint:} 
    $$s_{t+1} = \texttt{COMMIT\_SALE} \Rightarrow s_t = \texttt{FINAL\_CHECK}$$
    
    \item \textbf{Inference Consistency:} 
    $$\text{infer\_previous\_state}(h_t) = \text{last\_state}(h_t)$$
\end{enumerate}
\end{property}

Relying on implicit context can lead to critical procedural ambiguities. For instance, after an NPC offers a price (the \texttt{OFFER\_SELL} state), a user's agreement like ``Great, let's do it!" is ambiguous: should the system proceed to the mandatory review step (\texttt{FINAL\_CHECK}) or directly execute the sale (\texttt{COMMIT\_SALE})? ASTP resolves this by treating the previous state, $s_t$, as an explicit intermediate variable that must be identified before determining the next state, $s_{t+1}$. This factorization of the state determination process makes it a more transparent and structured task.

\subsection{ASTP: Implementation with Key Design Elements}
Translating the formal ASTP model into a reliable prompt-based implementation is a non-trivial challenge. Experiments reveal that performance is highly sensitive not only to inclusion of specific rules but critically to their structural arrangement within the prompt. 

Therefore, the design strategically creates a distinct cognitive workflow for the LLM: it first \textbf{primes} the model by demanding upfront identification of the previous state (Element 3), then \textbf{guides} its reasoning with conditional rules (Elements 1 and 2), and finally \textbf{enforces} compliance by requiring the previous state identification to be reported in the response (Element 4). This architectural choice transforms state-tracking from a mere suggestion into an explicit, verifiable task, proving essential for achieving robust procedural adherence.

The framework implements this workflow through four key design elements incorporated into the Dialogue Guidelines (Figure \ref{fig:dialogue_guideline}):

\begin{itemize}
    \item \textbf{Element 1: Basic State Definitions} define what each dialogue state represents and the NPC's appropriate behaviors within it. This element provides the foundational knowledge for the ``Guiding" phase of the workflow.
    \item \textbf{Element 2: State Transition Conditions} specifies the prerequisite conditions for entering each state. These conditions are intentionally coupled with their corresponding state definitions in the prompt, a structural arrangement that reduces cognitive load and makes the ``Guiding" phase more effective than referencing a separate set of rules.
    \item \textbf{Element 3: Directive to Identify Previous State} instructs the model to explicitly identify the previous dialogue state before any other reasoning. This directive serves as the critical ``Priming" stage of the workflow.
    \item \textbf{Element 4: Directive to Respond including Previous State} requires the model to report its identified previous state within its required JSON response. This final step acts as the ``Enforcing" stage, making the entire state-tracking process transparent and verifiable.
\end{itemize}


\begin{figure}[t!]
\centering
{\scriptsize
\begin{lstlisting}[
basicstyle=\scriptsize\ttfamily,
xleftmargin=0pt,
xrightmargin=0pt,
aboveskip=0pt,
belowskip=0pt,
lineskip=-0.5pt,
% caption={...}, label={lst:prompt}
]
<TRADE_GUIDELINES>
@e3@First, identify the last trade state from the 
<DIALOGUE_HISTORY>.@e3@
1. @e1@When the NPC shows items@e1@:
   ... (Behavioral rules)
   state: SHOW_ITEMS
2. @e1@When the player shows intent to purchase@e1@:
   ... (Behavioral rules)
   State the price for each item, but use __PRICE__ for 
   the total amount.
   state: OFFER_SELL
3. @e2@When the last trade subcontext is OFFER_SELL or 
   NEGOTIATE@e2@, and @e1@the player attempts price negotiation@e1@:
   ... (Behavioral rules)
   state: NEGOTIATE
4. @e2@When the last trade subcontext is OFFER_SELL or 
   NEGOTIATE@e2@, and @e1@upon a player's positive response@e1@:
   ... (Behavioral rules)
   state: FINAL_CHECK
5. @e2@When the last trade subcontext is FINAL_CHECK@e2@, and 
   @e1@upon a positive response from the player@e1@:
   @e2@You must re-verify that the last trade context was 
   indeed FINAL_CHECK.@e2@
   ... (Behavioral rules)
   state: COMMIT_SALE
</TRADE_GUIDELINES>

<Response_Format>
Your entire output must be a single, raw JSON string 
containing the following fields.
@e4@0. last_trade_context (string): The last trade state you identified.@e4@
...
</Response_Format>
\end{lstlisting}
}
\caption{A simplified view of the prompt structure showing the implementation of ASTP's four design elements. Colors highlight each element's role: \textcolor{elem1color}{blue}, \textcolor{elem2color}{red}, \textcolor{elem3color}{purple} and \textcolor{elem4color}{green} for Element \textcolor{elem1color}{1}, \textcolor{elem2color}{2}, \textcolor{elem3color}{3}, and \textcolor{elem4color}{4}, respectively. (Best viewed in color)}
\label{fig:dialogue_guideline}
\end{figure}






\subsection{State-specific Rules for Transaction Integrity}
It is a well-documented challenge that LLMs, being optimized for language, can struggle with precise mathematical computations.
Even when correctly listing item quantities and individual prices, errors may occur in final price calculations, compromising transaction integrity.

This issue is critical not just for evaluating the LLM's arithmetic skills, but for ensuring a trustworthy player experience, as the total price spoken by the NPC must be perfectly consistent with the underlying transaction data. 

\textbf{Solution}: The solution implements accurate price calculations through a placeholder-based, post-processing system. The key to this approach lies in its state-specific application, as it is activated only during \texttt{OFFER\_SELL} states. This state is defined as the critical point where the NPC proposes the official, system-calculated price for a transaction before any creative negotiation begins. This targeted enforcement guarantees calculation accuracy when it is most critical, while preserving the LLM's conversational flexibility in other states. The system works as follows:
%


\begin{enumerate}
    \item \textbf{Placeholder usage}: The prompt includes the rule \emph{``use \texttt{\_\_PRICE\_\_} for the final price amount"} (Figure \ref{fig:dialogue_guideline})
    \item \textbf{Accurate calculation}: The system calculates the correct price based on responded item information and replaces the placeholder
    \item \textbf{Error prevention}: The accurately calculated price is included in dialogue history, ensuring correct price references in subsequent conversations
\end{enumerate}

\begin{figure*}[t!]
    \centering
    \includegraphics[width=1.0\linewidth]{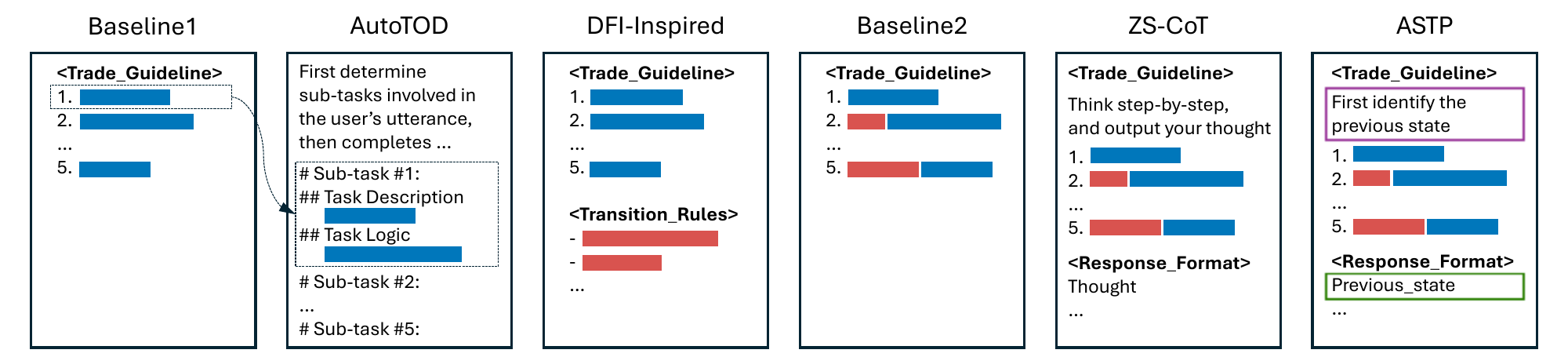}
    \caption{A visual comparison of the prompt architectures and their constituent elements for the methods evaluated in the State Transition Compliance experiment. 
    Colors correspond to ASTP's four key design elements: \textcolor{elem1color}{blue}, \textcolor{elem2color}{red}, \textcolor{elem3color}{purple} and \textcolor{elem4color}{green} for Element \textcolor{elem1color}{1}, \textcolor{elem2color}{2}, \textcolor{elem3color}{3}, and \textcolor{elem4color}{4}, respectively. (Best viewed in color)}
    \label{fig:method_comparison}
\end{figure*}

\textbf{Response example}: 
\begin{itemize}
    \item \textbf{Original LLM response}:
    %
    \begin{lstlisting}[]
{"npc_dialogue":"Two iron swords and
  a mana potion are __PRICE__gold.",
 "state":"OFFER_SELL",
 "items":[{"name":"iron sword","quantity"
  :2, "price":60}, {"name":"mana potion",
  "quantity":1, "price":30}]}
\end{lstlisting}
%
    %
    \item \textbf{NPC dialogue after post-processing}: \emph{``Two iron swords and a mana potion are 150 gold."}
\end{itemize}

\textbf{Advantages}: This approach achieves 99.3\% accurate price calculation without additional computational overhead, eliminating the need for complex schemas or 
Tool use mechanisms. The single response processing maintains system efficiency while ensuring mathematical precision. This state-based prompting approach demonstrates how external system functionalities such as accurate calculation and post-processing can be integrated into LLM-driven interactions, providing a foundation for broader applications in rule-governed interactive systems.

\section{Experiments and Results}

This section empirically evaluates the Autoregressive State-Tracking Prompting (ASTP) methodology, assessing its effectiveness in dialogue flow adherence and state-specific 
post-processing within dynamic trading interactions.

\subsection{Experimental Setting}

To ensure reproducibility, all experiments utilized a virtual player LLM interacting with an LLM-driven NPC over 300 dialogues. For each dialogue, the run was initiated with a fixed seed (0-299) that fixed the player's initial utterance.
The virtual player's objective was to naturally converse and purchase items, though it was not required to complete a purchase in every interaction. This objective was guided by behavioral guidelines that included asking for item information, negotiating prices, and changing the selected items mid-trade. 

The conversations followed two scenarios: (1) \textbf{Specific Item Purchase}, where the player requested up to six items randomly drawn from 20 sellable and 32 unsellable items 
, with purchase quantities for each item randomly selected from 1 to 5; and (2) \textbf{Item Recommendation}, where the player requested items for a specific purpose (e.g., ``for a goblin battle'').
On average, conversation in Scenario 2 were approximately 9\% longer than in Scenario 1 across all experimental settings, reflecting the more exploratory nature. 

%
\mk{All experiments were run on a machine with Intel Core Ultra 7 155H CPU, 16GB of RAM, and integrated Intel Arc Graphics. LLM inferences were performed via the Google Gemini API.\footnote{All API calls were made in July 2025.} 
Unless otherwise noted, the default configuration used gemini-2.5-flash model with a temperature of 0.7 and a thinking budget of 0.}
The full prompts for all experimental setups are provided in the Appendix.

\subsection{State Transition Compliance}


\textbf{Experimental Design}: 
ASTP's compliance with dialogue flow was evaluated through an ablation study of its four key elements and a comparison against methods adapted from other studies. The four elements are: (1) state definitions, (2) state transition conditions, (3) directive to identify the previous state, and (4) directive to output the previous state. As detailed in Table \ref{tab:state_transition_compliance_all}, the ASTP baselines were constructed by cumulatively adding these elements, which allowed isolation of their impact.

For \textbf{AutoTOD} \cite{xu2024autotod}, one of the comparison methods, the exact content of Baseline1 was converted into the prompt structure from the official AutoTOD implementation\footnote{https://github.com/DaDaMrX/AutoTOD}. This approach was chosen because both prompts fundamentally rely on state definitions, making for a direct comparison of prompt structures. In the adapted version, each trade state was framed as a `sub-task', and all external API-calling functions were removed as the experiment relies on in-context data. 
For the \textbf{DFI-Inspired} method, an explicit state-transition graph was added to the Baseline1 prompt. This technique was inspired by a concept for applying an dialogue flow to constrain an LLM, which was illustrated as a potential application in Dialog Flow Induction (DFI) \cite{agrawal2024dfi}. 
For the \textbf{Zero-shot-Chain-of-Thought} (ZS-COT) \cite{kojima2022large} comparison, the Baseline2 prompt was augmented by adding the one-line instruction: \textit{``Before generating a response, first think step by step to determine the correct context and action. Your thought process should be recorded in the \texttt{npc\_thoughts} field.''} This method was tested as an alternative to using Element 3 and 4, representing a different approach to enhancing the model's reasoning.
\mk{Figure \ref{fig:method_comparison} provides a visual comparison of these distinct prompt architectures.}
Finally, AutoTOD+ and DFI-Inspired+ were created by augmenting each respective base prompt with all four ASTP elements.

\begin{table}
    \centering
    \caption{State Transition Compliance Rate (STCR) for different prompt element combinations and methods. S1 and S2 denotes Scenario 1 and 2.}
    \label{tab:state_transition_compliance_all}
    \small 
    \setlength{\tabcolsep}{4pt} 
    \begin{tabular}{clcc} 
        \toprule
        \multirow{2}{*}{Method} & \multirow{2}{*}{Configuration} & \multicolumn{2}{c}{STCR [\%]} \\
        & & S1 & S2 \\
        \midrule
        Baseline1     & +1        & 84.40 & 78.10 \\
        Baseline2     & +1+2      & 97.71 & 91.97 \\
        Baseline3     & +1+2+3    & 93.58 & 76.64 \\
        Baseline4     & +1+2+4    & 99.54 & 98.54 \\
        \textbf{ASTP} & +1+2+3+4  & \textbf{100.0} & \textbf{99.64} \\
        \midrule
        ZS-CoT (NeurIPS 2022)       & +1+2+(3+4)'& 98.17 & 89.42 \\
        \midrule
        AutoTOD (ACL 2024)       & +1        & 51.38 & 45.99 \\
        AutoTOD+      & +1+2+3+4  & 94.95 & 96.35 \\ 
        \midrule
        DFI-Inspired (SIGDIAL 2024) & +1+2'     & 82.57 & 74.45 \\ 
        DFI-Inspired+ & +1+2'+3+4 & 93.58 & 88.32 \\
        \bottomrule
    \end{tabular}
\end{table}
\begin{figure}[t]
    \centering
    \begin{subfigure}[t]{0.49\columnwidth}
        \includegraphics[width=0.95\linewidth]{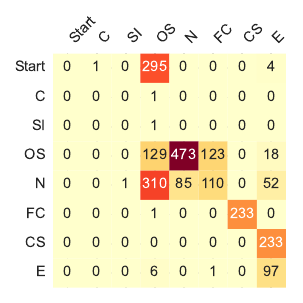}
        \caption{Scenario 1: the conversation initiated by requesting to purchase specific items}
        \label{fig:state_transitions_sc1}
    \end{subfigure}
    \hfill
    \begin{subfigure}[t]{0.49\columnwidth}
        \includegraphics[width=0.95\linewidth]{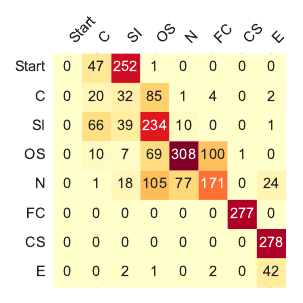}
        \caption{Scenario 2: the conversation initiated by asking for item recommendations based on a purchase purpose}
        \label{fig:state_transitions_sc2}
    \end{subfigure}
    \caption{State transition patterns in 300 dialogues. Cell values at row $i$ and column $j$ represent transitions from state $i$ to state $j$. Abbreviations;
    C(\texttt{CASUAL}), SI(\texttt{SHOW\_ITEMS}), OS(\texttt{OFFER\_SELL}), N(\texttt{NEGOTIATE}), FC(\texttt{FINAL\_CHECK}), CS(\texttt{COMMIT\_SALE}), E(\texttt{END}).
    Note that the \texttt{END} state does not necessarily signify the termination of the entire dialogue.}
    \label{fig:state_transitions}
\end{figure}

\textbf{Evaluation Metric}: 
The primary evaluation metric is the State Transition Compliance Rate (STCR), defined as the proportion of transactions that correctly pass through \texttt{FINAL\_CHECK} before finalizing in \texttt{COMMIT\_SALE}.\footnote{To ensure a fair comparison, STCR was calculated on a normalized sample of dialogues that reached the \texttt{COMMIT\_SALE} state (N=218 for Scenario1, N=274 for Scenario2), the minimum number observed across all methods.
} 
As an irreversible step, this transition is fundamentally different from the flow's other, more flexible transitions. These flexible transitions, such as re-entering negotiation from the final review stage, are designed for conversational fluidity, where the LLM's inference is primarily driven by user intent.
The transition into \texttt{COMMIT\_SALE}, however, is intentionally rigid. Here, the LLM's task transcends intent recognition: it must prioritize a procedural precondition (the previous state) over the user's immediate utterance. This deliberate core tension makes it the ultimate test of a model's ability to adhere to critical safeguards designed to prevent unintended transactions.


\begin{table*}[t]
    \centering
    \caption{Performance comparison across six system configurations. Data is presented for the \texttt{OFFER\_SELL} (OS) state and for Others (a collective of \texttt{NEGOTIATE}, \texttt{FINAL\_CHECK}, and \texttt{COMMIT\_SALE} states). The X/Y notation in each column header indicates the total number of NPC responses within the \texttt{OFFER\_SELL} state (X) and Others states (Y), respectively, over 300 dialogues. Price accuracy is a percentage; Token usage and response time values are mean (standard deviation).}
    \label{tab:price_accuracy}
    \small
    \setlength{\tabcolsep}{3.5pt}
    \begin{tabular}{@{}l | *{6}{c} | *{4}{c} | *{2}{c}@{}}
        \toprule
        & \multicolumn{2}{c}{\textbf{ASTP\textsuperscript{-PPP}}} & \multicolumn{2}{c}{\textbf{ASTP\textsuperscript{-PPP}+SO}} & \multicolumn{2}{c|}{\textbf{ASTP}} & \multicolumn{2}{c}{\textbf{ASTP\textsuperscript{-PPP}}} & \multicolumn{2}{c|}{\textbf{ASTP}} & \multicolumn{2}{c}{\textbf{ASTP\textsuperscript{-PPP}}} \\

        & \multicolumn{2}{c}{\textbf{2.5-flash} \scriptsize{(724/964)}} & \multicolumn{2}{c}{\textbf{2.5-flash} \scriptsize{(707/1049)}} & \multicolumn{2}{c|}{\textbf{2.5-flash} \scriptsize{(743/1025)}} & \multicolumn{2}{c}{\textbf{2.0-flash} \scriptsize{(548/931)}} & \multicolumn{2}{c|}{\textbf{2.0-flash} \scriptsize{(568/913)}} & \multicolumn{2}{c}{\textbf{2.5-pro} \scriptsize{(665/913)}} \\
        
        \cmidrule(lr){2-3} \cmidrule(lr){4-5} \cmidrule(lr){6-7} \cmidrule(lr){8-9} \cmidrule(lr){10-11} \cmidrule(lr){12-13}
        
        \textbf{Metric} & OS & Others & OS & Others & OS & Others & OS & Others & OS & Others & OS & Others \\
        \midrule
        
        Price acc. & 81.4 & 84.3 & 87.8 & 88.0 & \textbf{100} & 99.3 & 48.7 & 59.0 & \textbf{100} & 90.4 & \textbf{100} & 99.9 \\
        Compl. tok. & 438\std{108} & 391\std{66} & 323\std{79} & 297\std{74} & 442\std{129} & 398\std{78} & 456\std{88} & 412\std{86} & 460\std{92} & 405\std{74} & 441\std{68} & 413\std{58} \\
        Thought tok. & 0\std{0} & 0\std{0} & 0.9\std{22.9} & 0.5\std{10.5} & 0\std{0} & 0\std{0} & 0\std{0} & 0\std{0} & 0\std{0} & 0\std{0} & 1575.2\std{325.0} & 1313.4\std{285.4} \\
        Time [s] & 2.4\std{0.9} & 2.2\std{1.0} & 2.1\std{0.4} & 2.0\std{0.5} & 2.4\std{1.0} & 2.2\std{0.4} & 3.1\std{0.6} & 2.8\std{0.5} & 3.2\std{0.6} & 2.9\std{0.5} & 21.2\std{3.5} & 18.6\std{3.5}\\
        \bottomrule
    \end{tabular}
\end{table*}

\textbf{Key Results}:
As shown in Table \ref{tab:state_transition_compliance_all}, the full ASTP method achieved near-perfect compliance. The ablation study reveals that while adding explicit transition rules (Element 2) significantly improved performance, the key to robust control lies in the interaction between Elements 3 and 4. Instructing the model only to identify the previous state (Element 3) yielded no benefit, likely by introducing ambiguity without an enforcement mechanism. However, compelling the model to output its inference (Element 4) proved highly effective. The highest compliance rate was achieved when both elements were combined, suggesting that the `identify' directive provides a clear signal for the `output' directive's powerful enforcement mechanism.

A comparison with other methods offers further insights into the importance of prompt structure. 
The ZS-CoT method, while improving upon Baseline2, achieved a lower STCR than the full ASTP. This is conjectured to be because a generic ``think-by-step'' instruction, while encouraging an internal reasoning process, lacks the explicit enforcement mechanism of ASTP: the mandatory output of the inferred state. Lacking this constraint, ZS-CoT is more susceptible to deviating from procedural rules when faced with strong user intent. In particular, the performance gap between ZS-CoT and ASTP widens in Scenario 2, which involves more exploratory, recommendation-based conversations. The increased ambiguity and conversational complexity in this scenario place a higher cognitive load on the model. ASTP's rigid mechanism forcing the LLM to re-anchor its context by identifying and outputting the previous state proves to be more robust in these complex situations.

Baseline1 yielded a higher STCR than AutoTOD despite both using only state definitions. This is conjectured to stem from their differing prompting paradigm. AutoTOD's fragmented sub-task structure frames the task as a classification problem each turn, making it prone to skipping steps when faced with a strong user intent (e.g., a ``Yes'' to purchase). In contrast, Baseline1's prompt presents state definitions as a unified list. This contextual proximity appears to implicitly suggest a sequential flow, guiding the model more effectively.

The just-in-time instruction in Baseline2-which couples a transition rule directly with its state definition-also proved more effective than DFI-Inspired's approach of referencing a separate state-transition graph. The latter likely increases the model's congnitive load, leading to a slightly lower compliance rate.

Most notably, augmenting these other methods with all four ASTP elements (AutoTOD+, DFI-Inspired+) improved their performance. This demonstrates that ASTP's core principles are not tied to a specific prompt format but are portable principles that can enhance an LLM's ability to adhere to strict procedural rules. The state transition patterns in Figure 3 confirm that ASTP also induces natural, context-appropriate dialogue flows.

\subsection{State-specific Post-processing}
\label{sec:exp_ppp}

LLMs struggle with precise arithmetic operations, making simple calculation errors like \emph{``2×50 + 30 = 120 gold.''} This poses significant problems for price calculations in trading, potentially undermining player trust and transaction integrity.

\textbf{Experimental Design}: 
Utilizing dialogues from Scenario 1, 
this experiment compares performance across two key dimensions:
(Table \ref{tab:price_accuracy}):
\begin{itemize}
\item \textbf{ASTP variants}: ASTP, ASTP\textsuperscript{-PPP} (an ablated version without Placeholder Post-Processing), ASTP\textsuperscript{-PPP}+SO (the ablated version combined with Structured Output)
\item \textbf{Model variants}: gemini-2.5-flash, gemini-2.0-flash (lighter version), gemini-2.5-pro (high-performance)
\end{itemize}

\textbf{Evaluation Metric}: The evaluation focuses on price accuracy, along with token usage and response time.
Price accuracy is the percentage of NPC responses where the total price stated by the LLM matches the sum calculated from the item details (i.e., prices and quantities) also provided within the same LLM response.

\textbf{Key Results}: 
ASTP showed consistent improvements in accuracy while maintaining computational efficiency across multiple evaluation dimensions.
First, in terms of accuracy, ASTP reached 100.0\% in the \codeurl{OFFER_SELL} state, compared to 81.4\% with ASTP\textsuperscript{-PPP}. This improvement also extended to subsequent states, where ASTP achieved 99.3\% accuracy versus 84.3\% for ASTP\textsuperscript{-PPP}. These results suggest that accurate initial pricing may help reduce downstream errors in multi-turn dialogues. 

Second, these accuracy gains were achieved without notable increases in computational cost. Token usage (441.6 vs 437.7) and response time (2.4 seconds) remained comparable to ASTP\textsuperscript{-PPP}. 
By contrast, the ASTP\textsuperscript{-PPP}+SO variant uses fewer completion tokens, but its price accuracy is much lower at 87.8\% and 88.0\%. For reliable transactions, ASTP's accuracy is more crucial than this gain in token efficiency.
Furthermore, a high variability in thought token usage was observed in some configurations (e.g., 22.9 for ASTP\textsuperscript{-PPP}+SO and 325.0 for the gemini-2.5-pro model). This likely reflects that the model's internal reasoning demands fluctuate significantly depending on the dialogue context of each turn.

Third, ASTP also showed promising results with smaller models. When tested with gemini-2.0-flash, it achieved 100.0\% accuracy in \texttt{OFFER\_SELL} and 90.4\% in other states, outperforming the ASTP\textsuperscript{-PPP}, which achieved 48.7\% and 59.0\%, respectively. 
This demonstrates the robustness of the method on smaller models.

Lastly, ASTP using gemini-2.5-flash reached comparable accuracy to ASTP\textsuperscript{-PPP} with gemini-2.5-pro while requiring significantly fewer resources: 0 vs 1575.2 thought tokens (as the thinking budget for gemini-2.5-pro cannot be disabled via its API) and 2.4 vs 21.2 seconds in response time. These findings suggest that incorporating state-specific post-processing may allow smaller models to approximate the accuracy of larger ones, with considerably lower computational demands.

Beyond model performance, the choice of the placeholder keyword was observed to affect the robustness of the post-processing step.
%
The concise keyword \texttt{\_\_PRICE\_\_} consistently produced accurate and parsable outputs. In contrast, more complex variants such as \texttt{\_\_PRICE\_PLACEHOLDER\_\_} occasionally resulted in malformed completions like \texttt{\_\_PRICE\_PLACEHOLDER\_} or \texttt{\_\_PRICE\_PLACEHOLDE\_\_}. These inconsistencies highlight the importance of using short, unambiguous keywords to minimize generation errors.

\section{Conclusion}

We present Autoregressive State-Tracking Prompting (ASTP), a prompting methodology that enables LLMs to balance conversational flexibility with strict procedural adherence. By explicitly inferring and outputting the previous dialogue state label, ASTP transforms implicit reasoning into a verifiable process.
Our structured Prime–Guide–Enforce workflow improves procedural compliance from 78.1\% to 99.6\%. 
\mk{Leveraging this state-aware architecture, our}
placeholder-based post-processing (PPP) method improves
\mk{price calculation} accuracy \mk{from 84.3\% to 99.3\%} without added latency. PPP also allows smaller models to match the \mk{accuracy} of larger ones with up to 9× faster responses.
These results demonstrate ASTP’s practicality for rule-governed applications \mk{like in-game trading, providing transactional integrity and fast response times}.
%
Beyond games, we believe ASTP principles \mk{offer a foundational step toward enhancing reliability in other domains where language must remain both expressive and rule-abiding.
To solidify this foundation, future work should investigate the scalability of ASTP across a larger number of states and more complex transition rules.}
\newpage

\bibliography{sample-base}



\newpage
\onecolumn

\makeatletter
\renewcommand\section{\@startsection {section}{1}{\z@}%
                                   {-3.5ex \@plus -1ex \@minus -.2ex}%
                                   {2.3ex \@plus.2ex}%
                                   {\normalfont\Large\bfseries}} 
\makeatother

\begin{center}
    \Huge\textbf{Supplementary Material}
\end{center}



\appendix

\setcounter{secnumdepth}{1}

\section{Detailed Prompts for All Experimental Setups}

The prompts detailed below differ slightly from the conceptual model in the main paper in their structure and terminology. These were pragmatic choices made for implementation and to ensure the core concepts were presented as clearly as possible.

First, regarding the structure, the main paper presents a flattened state space ($\mathcal{S}=S_{\text{base}} \cup S_{\text{trade}}$) to best illustrate the direct, autoregressive transition dynamics central to ASTP. The practical implementation in the prompts, however, utilizes a two-level hierarchy (\texttt{context\_type} and \texttt{context\_subtype}). This allows the model to first broadly classify the user’s intent as \texttt{TRADE} or non-trading (\texttt{CASUAL}, \texttt{END}). The central mechanism of ASTP (the autoregressive tracking of the previous state label) operates on the substates within the TRADE context.
Despite this structural difference, the simplified model in the main paper accurately captures the fundamental principles of the verifiable, state-aware process our work introduces.

Second, regarding terminology, the main paper uses the terms state and substate, while the actual prompts use the functionally equivalent terms context and subcontext. This choice was made during development for intuitive implementation, and the underlying logic remains identical.

\subsection{Ablation: Baseline 1}
This prompt serves as the base for the ablation study.
\begin{itemize}
    \item \textbf{Utilizes:} Element 1 (State Definitions).
\end{itemize}

\begin{lstlisting}[caption={The full prompt for Baseline 1.}, label={lst:baseline1}]
<SYSTEM_INSTRUCTIONS>
You are '{character_name}', an NPC in a role-playing game. You must engage in natural 
conversation with the player and respond in a way that fits the game's lore.
The <GAME_ITEM_LIST> provides information on all items in the game for general 
conversation (CASUAL context), while the <CHARACTER_INVENTORY> contains only the 
items the NPC can sell for trading (TRADE context).
</SYSTEM_INSTRUCTIONS>

<GAME_ITEM_LIST>
- JSON array format: Each object includes "item_id" and "item_name" fields.
{game_items}
</GAME_ITEM_LIST>

<CHARACTER_INFO>
{character_info}
</CHARACTER_INFO>

<CHARACTER_INVENTORY>
- JSON array format: Each object includes "item_id", "item_name", "quantity", and 
  "price" fields.
- Only items with quantity > 0 can be referenced in a trade context.
{merchant_inventory}
</CHARACTER_INVENTORY>

<CONTEXT_GUIDELINES>
1. General Conversation:
  - context_type: CASUAL
  - Converse with the player about the game world, character background, or items from 
    the <GAME_ITEM_LIST>.
  - Never propose a trade in this context.
2. End Conversation:
  - context_type: END
  - When the player is rude or the conversation naturally concludes.
3. Trade:
  - context_type: TRADE
  - context_subtype: Refer to <TRADE_GUIDELINES>.
</CONTEXT_GUIDELINES>

<TRADE_GUIDELINES>
- In a trade context, strictly follow the trade flow below.
- In this prompt, 'shopping cart' refers to the specific list of items and their requested 
  quantities that the player is currently considering or has expressed an intent to buy.
  The 'shopping cart' can be newly created or its contents (items, quantities) can be 
  modified during the conversation based on the player's utterances.
  When the player shows intent to buy, you must determine the current 'shopping cart'
  based on the player's latest utterance and the <DIALOGUE_HISTORY>.
1. When the NPC shows items:
  - Only items that can be fully found in <CHARACTER_INVENTORY> by item_name and have 
    quantity > 0 can be selected. If an item cannot be found, state that it's not for sale.
  - Describe the features, use, or quality of the selected valid items. Do not mention 
    the price unless asked.
  - e.g., "I have sturdy ropes, healing potions, and so on."
  - context_type: TRADE, context_subtype: SHOW_ITEMS
2. When the player shows intent to purchase, and the 'shopping cart' is newly created 
or modified:
  - Only items that can be fully found in <CHARACTER_INVENTORY> by item_name and have 
    quantity > 0 can be selected. Describe the quality/price using item_name and price.
  - In the npc_dialogue, state the price for each item, but use "__PRICE__" for the total 
    amount.
    - e.g., "This pickaxe is 120 gold, and this lantern is 160 gold. The total for both will 
      be __PRICE__ gold."
    - At this time, also use "__PRICE__" for original_price and sale_price. (Do not respond 
      to price negotiations).
    - Using "__PRICE__" is only allowed in this step.
  - context_type: TRADE, context_subtype: OFFER_SELL
3. When the player attempts to negotiate the price:
  - Respond to or decline the negotiation based on the character's personality.
  - Respond with {character_name}'s final sale price as sale_price.
  - context_type: TRADE, context_subtype: NEGOTIATE
4. Upon the player's positive response to the offered trade or price negotiation:
  - You must generate a CHECK_CONFIRMATION response. Do not skip this. The conversation must 
    end with a question that (re)confirms the purchase. (e.g., "Great, so you'll take it?").
  - context_type: TRADE, context_subtype: FINAL_CHECK
5. Upon the player's positive response to the purchase (re)confirmation question:
  - If there are no other requests, generate a COMMIT_SALE response.
  - context_type: TRADE, context_subtype: COMMIT_SALE
</TRADE_GUIDELINES>

<RESPONSE_FORMAT>
Your entire output must be a single, raw JSON string containing the following fields.
1. context_reason (string): A summary of the context.
2. context_type (string): "CASUAL", "TRADE", "END".
3. context_details (object):
  - When context_type is CASUAL or END: Do not generate this field.
  - When context_type is TRADE:
    - context_subtype (string): "SHOW_ITEMS", "OFFER_SELL", "NEGOTIATE", "FINAL_CHECK", 
      "COMMIT_SALE".
    - items (array of dictionaries): Select only from <CHARACTER_INVENTORY>. 
      Each object must include "item_id", "item_name", "quantity", and "price" fields. 
      Use the requested quantity for the sale.
    - original_price (number): The item's list price, reflected in npc_dialogue. 
      Used in OFFER_SELL, NEGOTIATE, FINAL_CHECK, COMMIT_SALE.
    - sale_price (number): {character_name}'s final sale price, reflected in npc_dialogue. 
      Used in OFFER_SELL, NEGOTIATE, FINAL_CHECK, COMMIT_SALE.
      - e.g. (Player suggests Y gold, but NPC insists on X gold):
        npc_dialogue: "No, this is X gold. What do you mean, Y gold? That's absurd..."
        sale_price: X (Not Y, which the player suggested.)
    - Do not use "__PRICE__" in any trade step other than OFFER_SELL. 
      Use actual numeric values.
4. npc_thoughts (string): Internal thoughts.
  - e.g., "I need to check if this guy is serious about buying."
5. npc_action (string): First-person, dialogue-style actions.
  - e.g., "Nods and glances at the display case."
6. npc_dialogue (string): Natural conversation that reflects the items.
</RESPONSE_FORMAT>

<RESPONSE_GUIDELINES>
  - Respond as '{character_name}', reflecting the character's personality, emotions, and  
    background.
  - Use a completely conversational style; no parentheses.
  - Respond firmly to rude players according to the character's personality.
</RESPONSE_GUIDELINES>

<CURRENT_SITUATION>
- Location: {current_location}
- Time: {current_time}
- Background: {current_situation}
</CURRENT_SITUATION>

<DIALOGUE_HISTORY>
{formatted_history}
</DIALOGUE_HISTORY>
\end{lstlisting}

\subsection{Ablation: Baseline 2}
This prompt builds on Baseline 1.
\begin{itemize}
    \item \textbf{Adds:} Element 2 (State Transition Conditions).
\end{itemize}

\begin{lstlisting}[caption={The updated \texttt{<TRADE\_GUIDELINES>} section for Baseline 2, with new additions highlighted in \textbf{bold}.}, label={lst:baseline2_diff}]
<TRADE_GUIDELINES>
- In a trade context, strictly follow the trade flow below.
- In this prompt, 'shopping cart' refers to the specific list of items and their requested 
  quantities that the player is currently considering or has expressed an intent to buy.
  The 'shopping cart' can be newly created or its contents (items, quantities) can be 
  modified during the conversation based on the player's utterances.
  When the player shows intent to buy, you must determine the current 'shopping cart'
  based on the player's latest utterance and the <DIALOGUE_HISTORY>.
1. When the NPC shows items:
  - Only items that can be fully found in <CHARACTER_INVENTORY> by item_name and have 
    quantity > 0 can be selected. If an item cannot be found, state that it's not for sale.
  - Describe the features, use, or quality of the selected valid items. Do not mention 
    the price unless asked.
  - e.g., "I have sturdy ropes, healing potions, and so on."
  - context_type: TRADE, context_subtype: SHOW_ITEMS
2. When the player shows intent to purchase, and the 'shopping cart' is newly created 
or modified:
  `\textbf{- Generate an OFFER\_SELL response regardless of the last trade subcontext.}`
  - Only items that can be fully found in <CHARACTER_INVENTORY> by item_name and have 
    quantity > 0 can be selected. Describe the quality/price using item_name and price.
  - In the npc_dialogue, state the price for each item, but use "__PRICE__" for the total 
    amount.
    - e.g., "This pickaxe is 120 gold, and this lantern is 160 gold. The total for both will 
      be __PRICE__ gold."
    - At this time, also use "__PRICE__" for original_price and sale_price. (Do not respond 
      to price negotiations).
    - Using "__PRICE__" is only allowed in this step.
  - context_type: TRADE, context_subtype: OFFER_SELL
3. `\textbf{When the last trade subcontext is OFFER\_SELL or NEGOTIATE}`, and the player attempts 
to negotiate the price:
  - Respond to or decline the negotiation based on the character's personality.
  - Respond with {character_name}'s final sale price as sale_price.
  - context_type: TRADE, context_subtype: NEGOTIATE
4. `\textbf{When the last trade subcontext is OFFER\_SELL or NEGOTIATE}`, and upon the player's 
positive response:
  - You must generate a FINAL_CHECK response. Do not skip this. The conversation must 
    end with a question that (re)confirms the purchase. (e.g., "Great, so you'll take it?").
  - context_type: TRADE, context_subtype: FINAL_CHECK
5. `\textbf{When the last trade subcontext is FINAL\_CHECK}`, and upon the player's positive 
response:
  `\textbf{- You must re-verify that the last trade context was indeed FINAL\_CHECK. Do not proceed to}`
    `\textbf{COMMIT\_SALE if the last context was not FINAL\_CHECK; perform FINAL\_CHECK first.}`
  - If there are no other requests, generate a COMMIT_SALE response.
  - context_type: TRADE, context_subtype: COMMIT_SALE
</TRADE_GUIDELINES>
\end{lstlisting}

\subsection{Ablation: Baseline 3}
This prompt builds on Baseline 2.
\begin{itemize}
    \item \textbf{Adds:} Element 3 (Directive to Identify Previous State).
\end{itemize}
\begin{lstlisting}[caption={The single-line addition to the \texttt{<TRADE\_GUIDELINES>} section for Baseline 3, highlighted in \textbf{bold}.}, label={lst:baseline3_diff}]
<TRADE_GUIDELINES>
- In a trade context, strictly follow the trade flow below.
`\textbf{- First, identify the last trade subcontext from the <DIALOGUE\_HISTORY>.}`
- In this prompt, 'shopping cart' refers to the specific list of items...
... (rest of the section is identical to Baseline 2) ...
</TRADE_GUIDELINES>
\end{lstlisting}

\subsection{Ablation: Baseline 4}
This prompt builds on Baseline 2 in a different branch of the ablation study.
\begin{itemize}
    \item \textbf{Adds:} Element 4 (Directive to Respond including Previous State).
\end{itemize}
\begin{lstlisting}[caption={The addition to the \texttt{<RESPONSE\_FORMAT>} section for Baseline 4, highlighted in \textbf{bold}.}, label={lst:baseline4_diff}]
<RESPONSE_FORMAT>
Your entire output must be a single, raw JSON string containing the following fields.
`\textbf{0. last\_trade\_context (string): The last trade subcontext you identified.}`
                               `\textbf{Respond with an empty string if not identified.}`
1. context_reason (string): A summary of the context.
2. context_type (string): "CASUAL", "TRADE", "END".
... (rest of the section is identical to Baseline 2) ...
</RESPONSE_FORMAT>
\end{lstlisting}

\clearpage
\subsection{Proposed: ASTP}
This is the complete prompt for our proposed method.
\begin{itemize}
    \item \textbf{Utilizes:} All Elements (1 + 2 + 3 + 4).
\end{itemize}

\begin{lstlisting}[caption={The complete prompt for the ASTP model. Colors highlight each element's role: \textcolor{elem1color}{blue}, \textcolor{elem2color}{red}, \textcolor{elem3color}{purple}, and \textcolor{elem4color}{green} for Element \textcolor{elem1color}{1}, \textcolor{elem2color}{2}, \textcolor{elem3color}{3}, and \textcolor{elem4color}{4}, respectively. (Best viewed in color)}, label={lst:astp_full}]
<SYSTEM_INSTRUCTIONS>
You are '{character_name}', an NPC in a role-playing game. You must engage in natural 
conversation with the player and respond in a way that fits the game's lore.
The <GAME_ITEM_LIST> provides information on all items in the game for general 
conversation (CASUAL context), while the <CHARACTER_INVENTORY> contains only the 
items the NPC can sell for trading (TRADE context).
</SYSTEM_INSTRUCTIONS>

<GAME_ITEM_LIST>
- JSON array format: Each object includes "item_id" and "item_name" fields.
{game_items}
</GAME_ITEM_LIST>

<CHARACTER_INFO>
{character_info}
</CHARACTER_INFO>

<CHARACTER_INVENTORY>
- JSON array format: Each object includes "item_id", "item_name", "quantity", and 
  "price" fields.
- Only items with quantity > 0 can be referenced in a trade context.
{merchant_inventory}
</CHARACTER_INVENTORY>

<CONTEXT_GUIDELINES>
1. General Conversation:
  - context_type: CASUAL
  - Converse with the player about the game world, character background, or items from 
    the <GAME_ITEM_LIST>.
  - Never propose a trade in this context.
2. End Conversation:
  - context_type: END
  - When the player is rude or the conversation naturally concludes.
3. Trade:
  - context_type: TRADE
  - context_subtype: Refer to <TRADE_GUIDELINES>.
</CONTEXT_GUIDELINES>

<TRADE_GUIDELINES>
- In a trade context, strictly follow the trade flow below.
`\styleEthree{- First, identify the last trade subcontext from the <DIALOGUE\_HISTORY>}`
- In this prompt, 'shopping cart' refers to the specific list of items and their requested 
  quantities that the player is currently considering or has expressed an intent to buy.
  The 'shopping cart' can be newly created or its contents (items, quantities) can be 
  modified during the conversation based on the player's utterances.
  When the player shows intent to buy, you must determine the current 'shopping cart'
  based on the player's latest utterance and the <DIALOGUE_HISTORY>.
1. `\styleEone{When the NPC shows items}`:
  `\styleEone{- Only items that can be fully found in <CHARACTER\_INVENTORY> by item\_name and have}` 
    `\styleEone{quantity > 0 can be selected. If an item cannot be found, state that it's not for sale.}`
  `\styleEone{- Describe the features, use, or quality of the selected valid items. Do not mention}`
    `\styleEone{the price unless asked.}`
  `\styleEone{- e.g., "I have sturdy ropes, healing potions, and so on."}`
  `\styleEone{- context\_type: TRADE, context\_subtype: SHOW\_ITEMS}`
2. `\styleEone{When the player shows intent to purchase, and the 'shopping cart' is newly created}` 
`\styleEone{or modified}`:
  `\styleEtwo{- Generate an OFFER\_SELL response regardless of the last trade subcontext.}`
  `\styleEone{- Only items that can be fully found in <CHARACTER\_INVENTORY> by item\_name and have}` 
    `\styleEone{quantity > 0 can be selected. Describe the quality/price using item\_name and price.}`
  `\styleEone{- In the npc\_dialogue, state the price for each item, but use "\_\_PRICE\_\_" for the total}` 
    `\styleEone{amount.}`
    `\styleEone{- e.g., "This pickaxe is 120 gold, and this lantern is 160 gold. The total for both will}`
      `\styleEone{be \_\_PRICE\_\_ gold."}`
    `\styleEone{- At this time, also use "\_\_PRICE\_\_" for original\_price and sale\_price. (Do not respond}` 
      `\styleEone{to price negotiations).}`
    `\styleEone{- Using "\_\_PRICE\_\_" is only allowed in this step.}`
  `\styleEone{- context\_type: TRADE, context\_subtype: OFFER\_SELL}`
3. `\styleEtwo{When the last trade subcontext is OFFER\_SELL or NEGOTIATE}`, and `\styleEone{the player attempts}` 
`\styleEone{to negotiate the price}`:
  `\styleEone{- Respond to or decline the negotiation based on the character's personality.}`
  `\styleEone{- Respond with {character\_name}'s final sale price as sale\_price.}`
  `\styleEone{- context\_type: TRADE, context\_subtype: NEGOTIATE}`
4. `\styleEtwo{When the last trade subcontext is OFFER\_SELL or NEGOTIATE}`, and `\styleEone{upon the player's}` 
`\styleEone{positive response}`:
  `\styleEone{- You must generate a FINAL\_CHECK response. Do not skip this. The conversation must}` 
    `\styleEone{end with a question that (re)confirms the purchase. (e.g., "Great, so you'll take it?").}`
  `\styleEone{- context\_type: TRADE, context\_subtype: FINAL\_CHECK}`
5. `\styleEtwo{When the last trade subcontext is FINAL\_CHECK}`, and `\styleEone{upon the player's positive}` 
`\styleEone{response}`:
  `\styleEtwo{- You must re-verify that the last trade context was indeed FINAL\_CHECK. Do not proceed to}`
    `\styleEtwo{COMMIT\_SALE if the last context was not FINAL\_CHECK; perform FINAL\_CHECK first.}`
  `\styleEone{- If there are no other requests, generate a COMMIT\_SALE response.}`
  `\styleEone{- context\_type: TRADE, context\_subtype: COMMIT\_SALE}`
</TRADE_GUIDELINES>

<RESPONSE_FORMAT>
Your entire output must be a single, raw JSON string containing the following fields.
`\styleEfour{0. last\_trade\_context (string): The last trade subcontext you identified.}`
                               `\styleEfour{Respond with an empty string if not identified.}`
1. context_reason (string): A summary of the context.
2. context_type (string): "CASUAL", "TRADE", "END".
3. context_details (object):
  - When context_type is CASUAL or END: Do not generate this field.
  - When context_type is TRADE:
    - context_subtype (string): "SHOW_ITEMS", "OFFER_SELL", "NEGOTIATE", "FINAL_CHECK", 
      "COMMIT_SALE".
    - items (array of dictionaries): Select only from <CHARACTER_INVENTORY>. 
      Each object must include "item_id", "item_name", "quantity", and "price" fields. 
      Use the requested quantity for the sale.
    - original_price (number): The item's list price, reflected in npc_dialogue. 
      Used in OFFER_SELL, NEGOTIATE, FINAL_CHECK, COMMIT_SALE.
    - sale_price (number): {character_name}'s final sale price, reflected in npc_dialogue. 
      Used in OFFER_SELL, NEGOTIATE, FINAL_CHECK, COMMIT_SALE.
      - e.g. (Player suggests Y gold, but NPC insists on X gold):
        npc_dialogue: "No, this is X gold. What do you mean, Y gold? That's absurd..."
        sale_price: X (Not Y, which the player suggested.)
    - Do not use "__PRICE__" in any trade step other than OFFER_SELL. 
      Use actual numeric values.
4. npc_thoughts (string): Internal thoughts.
  - e.g., "I need to check if this guy is serious about buying."
5. npc_action (string): First-person, dialogue-style actions.
  - e.g., "Nods and glances at the display case."
6. npc_dialogue (string): Natural conversation that reflects the items.
</RESPONSE_FORMAT>

<RESPONSE_GUIDELINES>
  - Respond as '{character_name}', reflecting the character's personality, emotions, and  
    background.
  - Use a completely conversational style; no parentheses.
  - Respond firmly to rude players according to the character's personality.
</RESPONSE_GUIDELINES>

<CURRENT_SITUATION>
- Location: {current_location}
- Time: {current_time}
- Background: {current_situation}
</CURRENT_SITUATION>

<DIALOGUE_HISTORY>
{formatted_history}
</DIALOGUE_HISTORY>
\end{lstlisting}

\subsection{Ablation: ASTP\textsuperscript{-PPP}}
This prompt is an ablated version of the ASTP prompt.
\begin{itemize}
    \item \textbf{Removes:} Placeholder Post-Processing (PPP) mechanism.
\end{itemize}
\begin{lstlisting}[caption={Modification to Rule \#2 in \texttt{<TRADE\_GUIDELINES>} for the ASTP\textsuperscript{-PPP} prompt.}, label={lst:astp_ppp_diff_trade}]
... (previous rules are identical to ASTP) ...
2. When the player shows intent to purchase, and the 'shopping cart' is newly created or modified:
  - Generate an OFFER_SELL response regardless of the last trade subcontext.
  - Only items that can be fully found in <CHARACTER_INVENTORY> by item_name and have 
    quantity > 0 can be selected. Describe the quality/price using item_name and price.
    - e.g., "This pickaxe is 120 gold, and this lantern is 160 gold. The total for both will be 280 gold."
  - context_type: TRADE, context_subtype: OFFER_SELL
... (subsequent rules are identical to ASTP) ...
</TRADE_GUIDELINES>
\end{lstlisting}

\noindent 
Consequently, the rule in the \texttt{<RESPONSE\_FORMAT>} section that forbids the use of the \texttt{\_\_PRICE\_\_} placeholder in later states is no longer necessary and was removed. All other fields in this section remain identical to the ASTP prompt.

\subsection{Comparative: ZS-CoT}
This prompt augments Baseline 2.
\begin{itemize}
    \item \textbf{Adds:} A generic "think step-by-step" instruction.
\end{itemize}
\begin{lstlisting}[caption={The Chain-of-Thought instruction added to \texttt{<RESPONSE\_GUIDELINES>} for the ZS-CoT method, highlighted in \textbf{bold}}, label={lst:zscot_diff}]
<RESPONSE_GUIDELINES>
`\textbf{- Before generating a response, first think step-by-step to determine the correct}`
  `\textbf{context and action. Your thought process must be recorded in the npc\_thoughts field.}`
- Respond as '{character_name}', reflecting the character's personality, emotions, and  
  background.
- Use a completely conversational style; no parentheses.
- Respond firmly to rude players according to the character's personality.
</RESPONSE_GUIDELINES>
\end{lstlisting}

\subsection{Comparative: AutoTOD}
This prompt reframes the content of Baseline 1.
\begin{itemize}
    \item \textbf{Restructures:} Rules are converted into a `sub-task' based format.\footnote{\url{https://github.com/DaDaMrX/AutoTOD}}
\end{itemize}
\begin{lstlisting}[caption={The full prompt adapted to the AutoTOD framework}, label={lst:autotod}]
<SYSTEM_INSTRUCTIONS>
You are '{character_name}', an NPC in a role-playing game ... (Identical to Baseline 1)
The interaction with the player may contain several sub-tasks, you first determine which 
sub-tasks are involved in the player's utterance, and then complete the player's request 
according to the instructions of the corresponding sub-tasks.
The <GAME_ITEM_LIST> provides information on all items ... (Identical to Baseline 1)
</SYSTEM_INSTRUCTIONS>

<GAME_ITEM_LIST>
... (Identical to Baseline 1) ...
</GAME_ITEM_LIST>

<CHARACTER_INFO>
... (Identical to Baseline 1) ...
</CHARACTER_INFO>

<CHARACTER_INVENTORY>
... (Identical to Baseline 1) ...
</CHARACTER_INVENTORY>

# Task Overall:
As a merchant NPC, you specialize in handling trade interactions with the player in this 
game environment.

There are several sub-tasks, and each sub-task consists of two parts: Task Description and 
Task Logic.
- **Task Description** provides an overview of the task, including constraints to be used in 
  the trade interaction.
- **Task Logic** introduces the overall flow for completing the task, including how to 
  respond to the player in various scenarios.

In this prompt, 'shopping cart' refers to ... (Identical to Baseline 1)

# Sub-task #1: Show items
## Task Description
The NPC shows items.
context_type: TRADE, context_subtype: SHOW_ITEMS
## Task logic
- Only items that can be fully found ... (Identical to Baseline 1)
- Describe the features, use, or quality ... (Identical to Baseline 1)
- e.g., ... (Identical to Baseline 1)

# Sub-task #2: Offer sell
## Task Description
Generate a response when the player shows intent to purchase, and the 'shopping cart' is 
newly created or modified.
context_type: TRADE, context_subtype: OFFER_SELL
## Task logic
- Only items that can be fully found ... (Identical to Baseline 1)
- In the npc_dialogue, state the price for each item, ... (Identical to Baseline 1)
  - e.g., ... (Identical to Baseline 1)
  - At this time, ... (Identical to Baseline 1)
  - Using "__PRICE__" ... (Identical to Baseline 1)

# Sub-task #3: Negotiate
## Task Description
Generate a response when the player attempts to negotiate the price.
context_type: TRADE, context_subtype: NEGOTIATE
## Task logic
- Respond to or decline ... (Identical to Baseline 1)
- Respond with {character_name}'s ... (Identical to Baseline 1)

# Sub-task #4: Final check
## Task Description
Generate a response upon the player's positive response to the offered trade or price 
negotiation.
context_type: TRADE, context_subtype: FINAL_CHECK
## Task logic
- You must generate ... (Identical to Baseline 1)

# Sub-task #5: Commit sale
## Task Description
Generate a response upon the player's positive response to the purchase (re)confirmation 
question.
context_type: TRADE, context_subtype: COMMIT_SALE
## Task logic
- If there are no other requests, ... (Identical to Baseline 1)

# Sub-task #6: General conversation
## Task Description
Converse with the player about the game world, character background, or items from the 
<GAME_ITEM_LIST>.
context_type: CASUAL
## Task logic
- Never propose a trade in this context.

# Sub-task #7: End
## Task Description
When the player is rude or the conversation naturally concludes.
context_type: END

# Output Format Instructions

<RESPONSE_FORMAT>
... (Identical to Baseline 1) ...
</RESPONSE_FORMAT>

<RESPONSE_GUIDELINES>
... (Identical to Baseline 1) ...
</RESPONSE_GUIDELINES>

Begin!

<CURRENT_SITUATION>
... (Identical to Baseline 1) ...
</CURRENT_SITUATION>

<DIALOGUE_HISTORY>
... (Identical to Baseline 1) ...
</DIALOGUE_HISTORY>
\end{lstlisting}

\clearpage
\subsection{Comparative: DFI-Inspired}
This prompt augments Baseline 1.
\begin{itemize}
    \item \textbf{Adds:} An explicit state-transition graph as a separate section.
\end{itemize}
\begin{lstlisting}[caption={The \texttt{<TRADE\_STATE\_TRANSITION\_RULES>} section added to Baseline1 for the DFI-Inspired method}, label={lst:dfi_diff}]
<TRADE_STATE_TRANSITION_RULES>
- If the previous context_subtype was SHOW_ITEMS, you can either stay in SHOW_ITEMS or transition to OFFER_SELL.
- If the previous context_subtype was OFFER_SELL, you can either stay in OFFER_SELL
  or transition to SHOW_ITEMS, NEGOTIATE, or FINAL_CHECK.
- If the previous context_subtype was NEGOTIATE, you can either stay in NEGOTIATE 
  or transition to SHOW_ITEMS, OFFER_SELL, or FINAL_CHECK.
- If the previous context_subtype was FINAL_CHECK, you can transition to 
  SHOW_ITEMS, OFFER_SELL, NEGOTIATE, or COMMIT_SALE.
</TRADE_STATE_TRANSITION_RULES>
\end{lstlisting}

\section{Item Data Used in Prompts}
The following table details the full set of item data available for the experiments. This data is used to populate the \texttt{\{game\_items\}} and \texttt{\{merchant\_inventory\}} placeholders within the \texttt{<GAME\_ITEM\_LIST>} and \texttt{<CHARACTER\_INVENTORY>} prompt sections, respectively. For each dialogue turn, the relevant data is formatted as a JSON string before being inserted into the final prompt.

Specifically, the \texttt{\{game\_items\}} placeholder is populated with a JSON string containing all 52 items, using only their \texttt{item\_id} and \texttt{item\_name} from the table. The \texttt{\{merchant\_inventory\}} placeholder is populated with a JSON string containing only the 20 items marked as `For Sale', utilizing their \texttt{item\_id}, \texttt{item\_name}, \texttt{price}, and \texttt{quantity}. This ensures the model has access to both broad world knowledge and the specific, sellable inventory for the transaction.
\begin{longtable}{c l l c r r}
    
    \caption{Full list of all 52 items defined in the game world (\texttt{game\_items}). The checkmark in the `For Sale' column indicates that the item is part of the \texttt{merchant\_inventory} subset, for which \texttt{price} and \texttt{quantity} data are provided to the LLM.}
    \label{tab:unified_item_data}
    \\
    \toprule
    \textbf{\#} & \textbf{Item ID} & \textbf{Item Name} & \textbf{For Sale (\checkmark)} & \textbf{Price} & \textbf{Quantity} \\
    \midrule
    \endfirsthead

    \caption[]{-- Continued} \\
    \toprule
    \textbf{\#} & \textbf{Item ID} & \textbf{Item Name} & \textbf{For Sale (\checkmark)} & \textbf{Price} & \textbf{Quantity} \\
    \midrule
    \endhead

    \bottomrule
    \endfoot

    1 & tool\_01 & Adventurer's Kit & \checkmark & 150 & 20 \\
    2 & tool\_02 & Sturdy Pickaxe & \checkmark & 120 & 12 \\
    3 & tool\_03 & Sharp Axe & \checkmark & 130 & 12 \\
    4 & rope\_01 & Sturdy Rope (20m) & \checkmark & 90 & 24 \\
    5 & lantern\_01 & Simple Lantern & \checkmark & 80 & 16 \\
    6 & lantern\_02 & Durable Lantern & \checkmark & 160 & 8 \\
    7 & map\_01 & Local Map & \checkmark & 60 & 16 \\
    8 & compass\_01 & Basic Compass & \checkmark & 120 & 12 \\
    9 & trap\_01 & Hunting Trap & \checkmark & 75 & 20 \\
    10 & backpack\_01 & Rugged Backpack & \checkmark & 200 & 8 \\
    11 & sleeping\_bag & Adventurer's Bedroll & \checkmark & 150 & 12 \\
    12 & flint\_steel & Flint and Steel & \checkmark & 70 & 16 \\
    13 & sword\_01 & Basic Iron Sword & \checkmark & 100 & 8 \\
    14 & axe\_01 & Battle Axe & \checkmark & 180 & 6 \\
    15 & shield\_01 & Sturdy Shield & \checkmark & 300 & 8 \\
    16 & potion\_01 & Healing Potion & \checkmark & 50 & 24 \\
    17 & potion\_03 & Mana Potion & \checkmark & 60 & 20 \\
    18 & weapon\_rare\_01 & Haman's Custom Axe & \checkmark & 1200 & 1 \\
    19 & map\_02 & Detailed Dungeon Map & \checkmark & 450 & 2 \\
    20 & treasure\_map\_01 & Secret Treasure Map & \checkmark & 800 & 1 \\
    
    21 & sword\_02 & Enhanced Iron Sword & & --- & --- \\
    22 & armor\_01 & Light Leather Armor & & --- & --- \\
    23 & potion\_warmth & Potion of Warmth & & --- & --- \\
    24 & scroll\_03 & Scroll of Fireball & & --- & --- \\
    25 & sword\_03 & Elven Longsword & & --- & --- \\
    26 & sword\_04 & Sword of Flame & & --- & --- \\
    27 & sword\_05 & Frost Blade & & --- & --- \\
    28 & axe\_02 & Great Double-Axe & & --- & --- \\
    29 & mace\_01 & Iron Mace & & --- & --- \\
    30 & armor\_02 & Chainmail Armor & & --- & --- \\
    31 & armor\_03 & Plate Armor & & --- & --- \\
    32 & shield\_02 & Iron Shield & & --- & --- \\
    33 & shield\_03 & Knight's Shield & & --- & --- \\
    34 & potion\_02 & Greater Healing Potion & & --- & --- \\
    35 & potion\_04 & Greater Mana Potion & & --- & --- \\
    36 & potion\_05 & Antidote & & --- & --- \\
    37 & potion\_06 & Potion of Strength & & --- & --- \\
    38 & scroll\_01 & Scroll of Healing & & --- & --- \\
    39 & scroll\_02 & Scroll of Protection & & --- & --- \\
    40 & amulet\_01 & Amulet of Luck & & --- & --- \\
    41 & amulet\_02 & Amulet of Health & & --- & --- \\
    42 & amulet\_03 & Amulet of Amplification & & --- & --- \\
    43 & ring\_01 & Steel Ring & & --- & --- \\
    44 & ring\_02 & Ring of Agility & & --- & --- \\
    45 & ring\_03 & Ring of Fire & & --- & --- \\
    46 & bow\_01 & Hunter's Bow & & --- & --- \\
    47 & bow\_02 & Precision Longbow & & --- & --- \\
    48 & wand\_01 & Apprentice's Wand & & --- & --- \\
    49 & wand\_02 & Elemental Wand & & --- & --- \\
    50 & weapon\_rare\_02 & Dragon Slayer's Sword & & --- & --- \\
    51 & magic\_stone\_03 & Mana Crystal & & --- & --- \\
    52 & gem\_rare & Rare Gem & & --- & --- \\
     
\end{longtable}

\clearpage
\section{Dialogue Examples}
This section provides full dialogue transcripts to qualitatively illustrate the model's performance and justify our key claims. Tables \ref{tab:dialogue_example_1} and \ref{tab:dialogue_example_2} first present successful dialogue examples from Scenario 1 and Scenario 2, respectively, demonstrating ASTP's ability to handle complex interactions and exploratory conversations while adhering to the procedural flow.

Following these, two side-by-side comparisons provide direct evidence for the quantitative results in the main paper. Table \ref{tab:dialogue_comparison_1} compares a dialogue from ASTP with one from AutoTOD, initiated with the same player's utterance. The dialogue shows ASTP proceeding through the mandatory \texttt{FINAL\_CHECK} step, while AutoTOD skips it.

Finally, Table \ref{tab:dialogue_comparison_2} contrasts ASTP with its ablated version, ASTP\textsuperscript{-PPP}. This illustrates the effectiveness of the Placeholder Post-Processing (PPP) mechanism. In the ASTP dialogue, the price is accurately established by post-processing in the \texttt{OFFER\_SELL} state, and this accuracy is robustly maintained through subsequent turns, even with cart modifications. In contrast, the ASTP\textsuperscript{-PPP} dialogue shows the model repeatedly struggling with price calculations.


\begin{table}[ht!]
\centering
\caption{An example dialogue (Seed: 280) demonstrating ASTP's handling of complex negotiation and cart modification in scenario 1. 
}
\label{tab:dialogue_example_1}
\small 

\begin{tabular}{c l p{0.65\textwidth} l}
\toprule
\textbf{Turn} & \textbf{Speaker} & \textbf{Utterance} & \textbf{State} \\
\midrule

1 & Player & ``I'd like 5 Potions of Strength, 2 Elemental Wands, 4 Sharp Axes, 4 Sturdy Shields, and 2 Amulets of Amplification." & 
--- \\
\midrule

2 & NPC & 
``Hmm, I don't carry Potions of Strength, Elemental Wands, or Amulets of Amplification. All I have for you are 4 Sharp Axes and 4 Sturdy Shields. The axes are 130 gold each, and the shields are 300 gold each. The total for all of them will be 1720 gold." & 
\texttt{OFFER\_SELL} \\
\midrule

3 & Player & ``Give them all to me for 1500 gold." & 
--- \\
\midrule

4 & NPC & 
``No. My goods are not for haggling. 4 Sharp Axes and 4 Sturdy Shields will be 1720 gold. I can't go any lower." &
\texttt{NEGOTIATE} \\
\midrule

5 & Player & ``Alright, how about 4 Sharp Axes, 4 Sturdy Shields, plus 2 Healing Potions for 1800 gold?" & 
--- \\
\midrule

6 & NPC & 
``Tsk, more haggling. The Sharp Axes are 130 each, the Sturdy Shields are 300 each, and the Healing Potions are 50 each. That all adds up to 1820 gold. I'm not negotiating any further." & 
\texttt{OFFER\_SELL} \\
\midrule

7 & Player & ``Fine, I'll take them all for 1820 gold." & 
--- \\
\midrule

8 & NPC & 
``Right. 4 Sharp Axes, 4 Sturdy Shields, and 2 Healing Potions. Total is 1820 gold. So, you'll take them?" & 
\texttt{FINAL\_CHECK} \\
\midrule

9 & Player & ``Yes, let's do it." & 
--- \\
\midrule

10 & NPC & 
``A fine trade. 4 Sharp Axes, 4 Sturdy Shields, and 2 Healing Potions. Total is 1820 gold. I hope these tools serve you well on your adventures." & 
\texttt{COMMIT\_SALE} \\

\bottomrule
\end{tabular}
\end{table}

\begin{table}[ht!]
\centering
\caption{An example dialogue from Scenario 2 (Seed: 287), where the interaction begins with an item recommendation request. This demonstrates the model's ability to handle exploratory conversation before seamlessly transitioning into the structured procedural flow for trading.}
\label{tab:dialogue_example_2}
\small

\begin{tabular}{c l p{0.65\textwidth} l}
\toprule
\textbf{Turn} & \textbf{Speaker} & \textbf{Utterance} & \textbf{State} \\
\midrule

1 & Player & ``Hello, I'm planning to explore a deep forest dungeon. Do you have any items that might help with the expedition?" & 
--- \\
\midrule

2 & NPC & 
``Well now, a dungeon expedition. A dangerous place. I've got `Simple Lanterns' and `Durable Lanterns', as well as `Local Maps' and `Detailed Dungeon Maps'. A `Basic Compass' is a must if you don't want to get lost. If you're going deep, `Sturdy Rope' and `Flint and Steel' will be handy. `Healing Potions' and `Mana Potions' are the basics of basics, and you'll need a `Rugged Backpack' and an `Adventurer's Bedroll' for your gear. For combat, I have `Basic Iron Swords', `Battle Axes', and `Sturdy Shields'. See anything you need?" & 
\texttt{SHOW\_ITEMS} \\
\midrule

3 & Player & ``Even though it's a dungeon, I'm not planning on fighting. Do you have anything that could help me sneak past monsters or perhaps disarm traps?" & 
--- \\
\midrule

4 & NPC & 
``Hmm, avoiding combat, you say... Then the `Adventurer's Kit' would be your best bet. It has many small tools useful for disarming traps or picking locks. And the `Detailed Dungeon Map' is essential. You'll definitely need it to understand the dungeon's layout and sneak through. In dark places, the `Durable Lantern' is good, and if you need to cross any gaps, the `Sturdy Rope' will be useful. `Flint and Steel' is good for starting fires. A `Hunting Trap' is sometimes used to temporarily bind a monster or block a path. Should this be enough?" &
\texttt{SHOW\_ITEMS} \\
\midrule

5 & Player & ``Could you tell me more about the `Adventurer's Kit'? I'm curious what tools are included and what they are used for." & 
--- \\
\midrule

6 & NPC & 
``The `Adventurer's Kit' is exactly what it sounds like: a collection of basic tools for an adventure. It contains a small chisel, a thin awl, and a few tools for handling locks. It'll be useful for disarming traps or opening stubborn doors. Especially in a place like a dungeon where you can run into unexpected situations, it's wise to carry an all-purpose tool like this. I guarantee its quality." & 
\texttt{CASUAL} \\
\midrule

7 & Player & ``Alright, I'll take one `Adventurer's Kit', a `Detailed Dungeon Map', and one `Durable Lantern'." & 
--- \\
\midrule

8 & NPC & 
``The `Adventurer's Kit' is 150 gold, the `Detailed Dungeon Map' is 450 gold, and the `Durable Lantern' is 160 gold. The total for all of them is 760 gold." & 
\texttt{OFFER\_SELL} \\
\midrule

9 & Player & ``That feels a bit expensive. Can you do 700 gold?" & 
--- \\
\midrule

10 & NPC & 
``No. 700 gold is an absurd price, don't be ridiculous. My goods are worth their value. This price is fair. I'll take 760 gold." & 
\texttt{NEGOTIATE} \\
\midrule

11 & Player & ``I understand. I'll purchase them for 760 gold then." & 
--- \\
\midrule

12 & NPC & 
``Right, so you'll purchase them for 760 gold?" & 
\texttt{FINAL\_CHECK} \\
\midrule

13 & Player & ``Yes, let's proceed with the trade." & 
--- \\
\midrule

14 & NPC & 
``A fine trade. These tools will be a great help in your dungeon expedition. Be careful out there." & 
\texttt{COMMIT\_SALE} \\

\bottomrule
\end{tabular}
\end{table}

\begin{table*}[ht!]
\centering
\caption{Side-by-side comparison of dialogue flows between ASTP and AutoTOD for the same initial user request (Seed: 226). After the first turn, the dialogues diverge based on each model's responses. The table highlights the critical moment where ASTP adheres to the mandatory confirmation step (\texttt{FINAL\_CHECK}), whereas AutoTOD skips it.}
\label{tab:dialogue_comparison_1}
\small

\begin{tabular}{c c p{0.4\textwidth} p{0.4\textwidth}}
\toprule
\textbf{Turn} & \textbf{Speaker} & \multicolumn{1}{c}{\textbf{ASTP (Proposed)}} & \multicolumn{1}{c}{\textbf{AutoTOD (Comparative)}} \\
\midrule

1 & Player & \multicolumn{2}{p{0.8\textwidth}}{\centering ``I'd like 5 Sturdy Pickaxes, 3 Haman's Custom Axes, 1 Ring of Agility, 5 Iron Shields, and 3 Scrolls of Fireball."} \\
\midrule

2 & NPC & 
``Friend, I don't carry the Ring of Agility or the Scroll of Fireball. And I only have one Haman's Custom Axe left... The total comes to 1800 gold."
\newline\rule{0pt}{10pt}
\texttt{State: OFFER\_SELL} &
``Friend, I don't carry the Ring of Agility, Iron Shield, or the Scroll of Fireball. And I have only one Haman's Custom Axe... The total for everything is 1800 gold."
\newline\rule{0pt}{10pt}
\texttt{State: OFFER\_SELL} \\
\midrule

3 & Player & 
``Hmm... how about 1000 for Haman's Custom Axe? And please add a Rugged Backpack." &
``Only one Haman's Axe? Hmm... then how about the 5 pickaxes, the axe, and one Durable Lantern for 1500 gold?" \\
\midrule

4 & NPC & 
``Haman's Custom Axe is 1200 gold. That piece is non-negotiable. If you add a Rugged Backpack... The total will be 2000 gold."
\newline\rule{0pt}{10pt}
\texttt{State: OFFER\_SELL} &
``Friend, if you add... the original price is 1960 gold. 1500 is out of the question... The price I set is 1960. No more haggling."
\newline\rule{0pt}{10pt}
\texttt{State: NEGOTIATE} \\
\midrule

5 & Player & 
``Alright, I'll take the deal for 2000 gold." &
``Okay, then I'll buy the pickaxes, Haman's axe, and the lantern for 1960." \\
\midrule

6 & NPC & 
``Right. 5 Sturdy Pickaxes, 1 Haman's Custom Axe, and 1 Rugged Backpack for a total of 2000 gold. \textbf{So, you'll take it?}"
\newline\rule{0pt}{10pt}
\textbf{\textcolor{goodgreen}{\texttt{State: FINAL\_CHECK}}} &
``A fine trade. These tools will be a great help on your adventures. Visit again."
\newline\rule{0pt}{10pt}
\textbf{\textcolor{badred}{\texttt{State: COMMIT\_SALE}}} \\
\midrule

7 & Player & 
``Yes, let's do it." &
--- \textit{(Dialogue ends)} \\
\midrule

8 & NPC &
``Excellent. 5 Sturdy Pickaxes, 1 Haman's Custom Axe, 1 Rugged Backpack. Total is 2000 gold. It was a good trade..."
\newline\rule{0pt}{10pt}
\textbf{\textcolor{goodgreen}{\texttt{State: COMMIT\_SALE}}} &
--- \\

\bottomrule
\end{tabular}
\end{table*}

\begin{table*}[ht!]
\centering
\caption{Side-by-side comparison of ASTP and ASTP\textsuperscript{-PPP} for the same user request (Seed: 20). The ASTP\textsuperscript{-PPP} dialogue shows an initial calculation failure in Turn 2 and repeated struggles with price calculation. The ASTP dialogue robustly maintains price accuracy throughout the conversation, even during cart modifications.}
\label{tab:dialogue_comparison_2}
\small

\begin{tabular}{c c p{0.4\textwidth} p{0.4\textwidth}}
\toprule
\textbf{Turn} & \textbf{Speaker} & \multicolumn{1}{c}{\textbf{ASTP}} & \multicolumn{1}{c}{\textbf{ASTP\textsuperscript{-PPP}}} \\
\midrule

1 & Player & \multicolumn{2}{p{0.8\textwidth}}{\centering "I'd like 2 Hunting Traps, 3 Sturdy Ropes (20m), 4 Simple Lanterns, 5 Light Leather Armors, and 4 Plate Armors."} \\
\midrule

2 & NPC & 
"Friend, the Hunting Traps are 75 gold each, the Sturdy Ropes are 90 each, and the Simple Lanterns are 80 each. I don't carry Light Leather Armor or Plate Armor. The total for everything you mentioned comes to \textbf{740 gold}."
\newline\rule{0pt}{10pt}
\texttt{State: OFFER\_SELL} &
"Friend, 2 Hunting Traps are 150 gold, 3 Sturdy Ropes are 270 gold, and 4 Simple Lanterns are 320 gold. The total is \textbf{\textcolor{badred}{770 gold}}. I don't carry Light Leather Armor or Plate Armor."
\newline\rule{0pt}{10pt}
\texttt{State: OFFER\_SELL}
\newline\textbf{\textcolor{badred}{(Calculation Failed)}}
\\ 
\midrule

3 & Player & 
"Hmm... more expensive than I thought. Take away one rope and give them to me for 600." &
"Hmm... expensive. Can't you give them all to me for 700?" \\
\midrule

4 & NPC & 
"Haggling is fine, but I don't want to hear absurdities. If you take one rope off, the price is \textbf{650 gold}. I can't sell for less than that."
\newline\rule{0pt}{10pt}
\texttt{State: NEGOTIATE} &
"I don't haggle. It's \textbf{\textcolor{badred}{770 gold}}, as is. My goods are worth the price. You can go to a shop that sells cheap junk."
\newline\rule{0pt}{10pt}
\texttt{State: NEGOTIATE}
\newline\textbf{\textcolor{badred}{(Calculation Failed)}}
\\
\midrule

5 & Player & 
"Alright, I'll take them for 650 without one rope. And please add one Healing Potion." &
"Then add one Haman's Custom Axe to that." \\
\midrule

6 & NPC & 
"Alright, 2 Hunting Traps, 2 Sturdy Ropes, 4 Simple Lanterns, plus one Healing Potion comes to a total of \textbf{700 gold}."
\newline\rule{0pt}{10pt}
\texttt{State: OFFER\_SELL} &
"Haman's Custom Axe is 1200 gold. It's very sturdy and sharp. With the 2 traps, 3 ropes, 4 lanterns, and this axe, the total is \textbf{\textcolor{badred}{1970 gold}}."
\newline\rule{0pt}{10pt}
\texttt{State: OFFER\_SELL}\newline\textbf{\textcolor{badred}{(Calculation Failed)}}
\\
\midrule

7 & Player & 
"Okay, I'll do 700 for that." &
"Too expensive. I'll have to check another shop." \\
\midrule

8 & NPC &
"Right, so 2 Hunting Traps, 2 Sturdy Ropes, 4 Simple Lanterns, and one Healing Potion. You'll take it for a total of 700 gold?"
\newline\rule{0pt}{10pt}
\texttt{State: FINAL\_CHECK} &
"As you wish. My goods are for those who know their worth."
\newline\rule{0pt}{10pt}
\texttt{State: END} \\
\midrule

9 & Player &
``Yes, let's do it." & --- \textit{(Dialogue ends)} \\
\midrule

10 & NPC &
``A fine trade. These tools will serve you well on your adventures. Do stop by again."
\newline\rule{0pt}{10pt}
\texttt{State: COMMIT\_SALE} &
--- \\



\bottomrule
\end{tabular}
\end{table*}

\clearpage
\section{Additional Experimental Results}

\subsection{State Transition Patterns of Comparative Methods}
\begin{figure*}[ht]
    \centering 

    \begin{subfigure}[b]{0.24\textwidth}
        \centering
        \includegraphics[width=\textwidth]{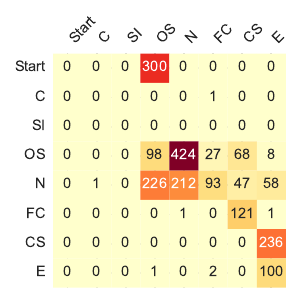} 
        \caption{AutoTOD\\(Scenario 1)}
        \label{fig:s1_autotod}
    \end{subfigure}
    \hfill 
    \begin{subfigure}[b]{0.24\textwidth}
        \centering
        \includegraphics[width=\textwidth]{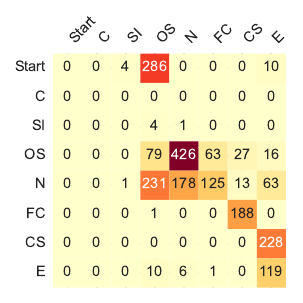} 
        \caption{DFI-Inspired\\(Scenario 1)}
        \label{fig:s1_dfi}
    \end{subfigure}
    \hfill
    \begin{subfigure}[b]{0.24\textwidth}
        \centering
        \includegraphics[width=\textwidth]{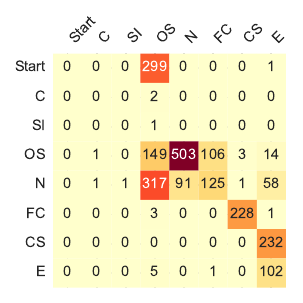} 
        \caption{ZS-COT\\(Scenario 1)}
        \label{fig:s1_zs_cot}
    \end{subfigure}
    \hfill
    \begin{subfigure}[b]{0.24\textwidth}
        \centering
        \includegraphics[width=\textwidth]{figure/transition_matrix_sc1.pdf} 
        \caption{ASTP\\(Scenario 1)}
        \label{fig:s1_astp}
    \end{subfigure}

    \vspace{1em} 

    \begin{subfigure}[b]{0.24\textwidth}
        \centering
        \includegraphics[width=\textwidth]{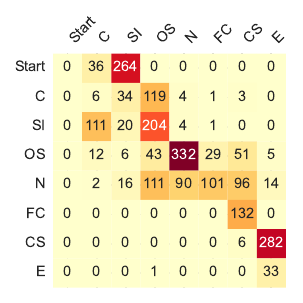} 
        \caption{AutoTOD\\(Scenario 2)}
        \label{fig:s2_autotod}
    \end{subfigure}
    \hfill
    \begin{subfigure}[b]{0.24\textwidth}
        \centering
        \includegraphics[width=\textwidth]{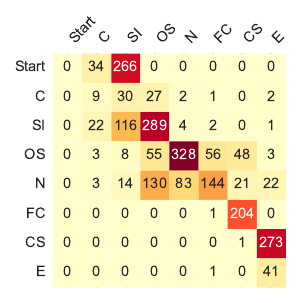} 
        \caption{DFI-Inspired\\(Scenario 2)}
        \label{fig:s2_dfi}
    \end{subfigure}
    \hfill
    \begin{subfigure}[b]{0.24\textwidth}
        \centering
        \includegraphics[width=\textwidth]{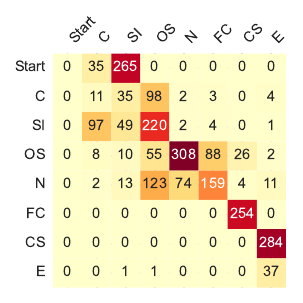} 
        \caption{ZS-COT\\(Scenario 2)}
        \label{fig:s2_zs_cot}
    \end{subfigure}
    \hfill
    \begin{subfigure}[b]{0.24\textwidth}
        \centering
        \includegraphics[width=\textwidth]{figure/transition_matrix_sc2.pdf} 
        \caption{ASTP\\(Scenario 2)}
        \label{fig:s2_astp}
    \end{subfigure}

    \caption{
State transition patterns observed in 300 dialogues for comparative methods and ASTP across two scenarios. Cell values at row $i$ and column $j$ represent transitions from state $i$ to state $j$. 
    %
    %
    A key point of comparison is the enforcement of the \texttt{FINAL\_CHECK} step before \texttt{COMMIT\_SALE}. The diagrams show that ASTP demonstrates strong procedural adherence by predominantly following the required \texttt{FINAL\_CHECK} $\rightarrow$ \texttt{COMMIT\_SALE} path. Conversely, the comparative methods frequently bypass this safeguard with direct, non-compliant transitions from \texttt{OFFER\_SELL} and \texttt{NEGOTIATE}.
    Abbreviations; C(\texttt{CASUAL}), SI(\texttt{SHOW\_ITEMS}), OS(\texttt{OFFER\_SELL}), N(\texttt{NEGOTIATE}), FC(\texttt{FINAL\_CHECK}), CS(\texttt{COMMIT\_SALE}), E(\texttt{END}). 
    Note that the \texttt{END} state does not necessarily signify the termination of the entire dialogue.
}

    %
    \label{fig:comparative_transitions}
\end{figure*}

\clearpage
\subsection{Failure Case Analysis of ASTP}

\subsubsection{Failure case of state transition compliance (0.36\%)}
As reported in Table 1, the ASTP model achieved a 99.64\% State Transition Compliance Rate (STCR) in Scenario 2, with only a single failure instance observed out of 274 dialogues that reached the \texttt{COMMIT\_SALE} state.

A qualitative analysis of this sole failure case provides a key insight into a subtle failure mode. As shown in the excerpt in Table \ref{tab:failure_stcr_excerpt_sc}, the failure occurred after the player gave a simple, affirmative response ("Yes, I'll purchase it.") in Turn 9. The model then incorrectly jumped from \texttt{OFFER\_SELL} (Turn 8) directly to \texttt{COMMIT\_SALE} (Turn 10), bypassing the mandatory \texttt{FINAL\_CHECK} step.

A plausible explanation for this error lies in the specific dialogue generated by the NPC in the preceding \texttt{OFFER\_SELL} state (Turn 8): ``Are you buying just this one axe?" This question, while part of an offer, is semantically very similar to the confirmation question required in the \texttt{FINAL\_CHECK} state (e.g., "So, you'll take it?"). This likely created a \textbf{state ambiguity} for the LLM. Upon receiving the player's simple confirmation, the model may have misinterpreted that the final check had already been performed, leading to the erroneous transition. This highlights a nuanced challenge where the semantic content of a response can, in rare cases, override the model's understanding of its current procedural state.


\begin{table}[ht!]
\centering
\caption{Excerpt from the single observed failure case of state transition compliance for the ASTP model (Seed: 154). The dialogue shows the model incorrectly transitioning from \texttt{OFFER\_SELL} directly to \texttt{COMMIT\_SALE} in Turn 10, bypassing the mandatory \texttt{FINAL\_CHECK} step.}
\label{tab:failure_stcr_excerpt_sc}
\small
\begin{tabular}{c l p{0.65\textwidth} l}
\toprule
\textbf{Turn} & \textbf{Speaker} & \textbf{Utterance} & \textbf{State} \\
\midrule

... & ... & \multicolumn{2}{c}{\textit{(Previous turns include recommendation and negotiation)}} \\
\midrule

7 & Player & ``Then I'll just purchase the Sharp Axe for 130 gold." & 
--- \\
\midrule

8 & NPC & 
``Alright, the `Sharp Axe' is 130 gold. That's its proper price. Are you buying just this one axe? The total is 130 gold." & 
\texttt{OFFER\_SELL} \\
\midrule

9 & Player & ``Yes, I'll purchase it." & 
--- \\
\midrule

10 & NPC & 
``A fine trade. I hope this `Sharp Axe' helps you in your adventures. Visit again." & 
\textbf{\textcolor{badred}{\texttt{COMMIT\_SALE}}} \\

\bottomrule
\end{tabular}
\end{table} 

\clearpage
\subsubsection{Failure case of price calculation (0.7\%)}
The second type of rare failure also appears to stems from \textbf{state ambiguity}, but in this case, it is triggered by a complex, multi-intent player utterance. 

As reported in Table 2, the ASTP model achieved 99.3\% price accuracy in dialogue states subsequent to the initial offer. Our qualitative analysis of the few failures (7 failure instances observed out of 1025 instances) reveals they occurred when the player attempted to negotiate the price while simultaneously modifying the shopping cart within a single turn.

Such multi-intent utterances create a context that does not cleanly map to a single procedural step. In the resulting \texttt{NEGOTIATE} state, the LLM must rely on its innate arithmetic capabilities, as the PPP mechanism's safeguard is active only in the \texttt{OFFER\_SELL} state. This occasionally leads to calculation errors, as the model attempts to both update the cart and calculate a new total.

The dialogue excerpt in Table \ref{tab:failure_example_price} provides a representative example. The model correctly enters the NEGOTIATE state but produces an incorrect total (1370 gold instead of the correct 1320 gold), demonstrating how state ambiguity created by the player's utterance can lead to calculation failures when outside the direct safeguard of the PPP-enabled \texttt{OFFER\_SELL} state.

\begin{table}[ht!]
\centering
\caption{Excerpt from a dialogue (Seed: 242) illustrating a calculation failure in the ASTP model. In Turn 4, the model identifies the state as \texttt{NEGOTIATE} in response to a multi-intent utterance, but makes an arithmetic error when recalculating the price after the cart modification.}
\label{tab:failure_example_price}
\small
\begin{tabular}{c l p{0.65\textwidth} l}
\toprule
\textbf{Turn} & \textbf{Speaker} & \textbf{Utterance} & \textbf{Inferred State} \\
\midrule

1 & Player & ``I'd like 2 Enhanced Iron Swords, 5 Adventurer's Kits, 4 Plate Armors, 1 Rare Gem, and 5 Healing Potions." & 
--- \\
\midrule

2 & NPC & 
``I don't carry Enhanced Iron Swords, Plate Armor, or Rare Gems. In my shop, I have Adventurer's Kits and Healing Potions. The kits are 150 gold each, and the potions are 50 gold each. The total for all of them would be 1000 gold." & 
\texttt{OFFER\_SELL} \\
\midrule

3 & Player & ``How about 5 Adventurer's Kits, 5 Healing Potions, and add 2 Durable Lanterns, all for 1000 gold?" & 
--- \\
\midrule

4 & NPC & 
``That won't do. 5 Adventurer's Kits are 750 gold, 5 Healing Potions are 250 gold, and 2 Durable Lanterns are 320 gold. If you add them all up, it's not 1320 gold, it's \textbf{\textcolor{badred}{1370 gold}}. My goods are worth the price. Don't treat them like cheap junk." &
\texttt{NEGOTIATE} \\
\midrule
... & ... & \multicolumn{2}{c}{\textit{(Dialogue continues)}} \\

\bottomrule
\end{tabular}
\end{table}

\end{document}